\documentclass[lettersize,journal,usenames,dvipsnames]{IEEEtran}

\usepackage{cite} 
\usepackage[acronym]{glossaries} 
\usepackage{booktabs} 
\usepackage{xspace} 
\usepackage{url} 
\usepackage[ruled,vlined,linesnumbered]{algorithm2e} 
\usepackage{amsthm} 
\usepackage{bm} 
\usepackage{MnSymbol}
\usepackage{nicefrac} 
\usepackage[inline]{enumitem} 
\usepackage{graphicx}
\usepackage{amsfonts} 
\usepackage{suffix} 
\usepackage{framed} 
\usepackage{xcolor} 
\usepackage[print-env=true]{scontents}
\usepackage{hyperref} 

\newtheorem{theorem}{Theorem}

\newtheorem{proposition}[theorem]{Proposition}

\usepackage[cal=boondoxo]{mathalfa} 

\renewcommand{\b}[1]{\bm{#1}} 
\renewcommand{\c}[1]{\mathcal{#1}} 
\newcommand{\bc}[1]{\b{\c{#1}}} 

\newcommand{\high}[1]{\colorlet{highprev}{.}\color{Black}#1\color{highprev}\xspace}

\renewenvironment{quote}{\begin{fquote}\itshape\advance\leftmargini -2.4em\begin{oldquote}}{\end{oldquote}\end{fquote}}

\newenvironment{fquote}
{%
	\MakeFramed {\advance\hsize-3\width \FrameRestore}
	}
	{\endMakeFramed}

\long\def\RC#1\par{\setlength{\parskip}{0.6\baselineskip}\setlength{\parindent}{0pt}\makebox[0pt][r]{\bf RC:\hspace{4mm}}\textbf{{#1}}\par} 
\WithSuffix\long\def\RC*#1\par{\textbf{{#1}}\par} 

\long\def\AR#1\par{\setlength{\parskip}{0.6\baselineskip}\setlength{\parindent}{0pt}\makebox[0pt][r]{AR:\hspace{10pt}}{#1}\par} 
\WithSuffix\long\def\AR*#1\par{{#1}\par} 

\newacronym{cnn}{CNN}{Convolutional Neural Networks}
\newacronym[longplural={Dynamical Systems}]{ds}{DS}{Dynamical System}
\newacronym{dmp}{DMP}{Dynamic Movement Primitive}
\newacronym[longplural={Degrees of Freedom}]{dof}{DoF}{Degree of Freedom}
\newacronym{dl}{DL}{Deep Learning}
\newacronym[longplural={Gausian Processes}]{gp}{GP}{Gaussian Process}
\newacronym[longplural={Graph Gausian Processes}]{ggp}{GGP}{Graph Gaussian Process}
\newacronym{gmm}{GMM}{Gaussian Mixture Model}
\newacronym{gpr}{GPR}{Gaussian Process Regression}
\newacronym{iil}{IIL}{Interactive Imitation Learning}
\newacronym{hmm}{HMM}{Hidden Markov Model}
\newacronym{hsmm}{HSMM}{Hidden Semi-Markov Model}
\newacronym{kt}{KT}{kinesthetic teaching}
\newacronym{kd}{KD}{kinesthetic demonstration}
\newacronym{lfd}{LfD}{Learning from Demonstrations}
\newacronym{lstm}{LSTM}{Long Short-Term Memory}
\newacronym{pca}{PCA}{Principal Component Analysis}
\newacronym{mp}{MP}{Movement Primitive}
\newacronym{nn}{NN}{Neural Network}
\newacronym{rl}{RL}{Reinforcement Learning}
\newacronym{rbf}{RBF}{Radial Basis Function}
\newacronym{simple}{SIMPLe}{Safe Interactive Movement Primitive Learning}
\newacronym{tphsmm}{TPHSMM}{Task Parameterized Hidden Semi-Markov Model}
\newacronym{tpgmm}{TPGMM}{Task Parameterized Gaussian Mixture Model}

\newcommand{\cartPoses}{\b{\xi}}
\newcommand{\cartPose}{\b{x}}
\newcommand{\timeReal}{t}
\newcommand{\timeBelief}{t^b}
\newcommand{\timestamps}{\b{\tau}}
\newcommand{\policy}{\pi}

\newcommand{\pPolicyState}{\cartPose}
\newcommand{\pPolicyGoal}{\cartPose_g}
\newcommand{\tPolicyState}{\timeReal}
\newcommand{\tPolicyGoal}{\cartPose_g}
\newcommand{\ptPolicyState}{\cartPose, \timeBelief}
\newcommand{\ptPolicyGoal}{\cartPose_g, \timeBelief_g}
\newcommand{\pPolicy}{\policy_{\pPolicyState}}
\newcommand{\tPolicy}{\policy_{\tPolicyState}}
\newcommand{\ptPolicy}{\policy_{\ptPolicyState}}

\newcommand{\covTrainTrain}{\bc{K}}
\newcommand{\covPointTrain}{\bc{k}_\star}
\newcommand{\covPointPoint}{\bc{k}}
\newcommand{\graphCovPointTrain}{\tilde{\bc{k}}_\star}
\newcommand{\gpState}{\bc{x}}
\newcommand{\gpStates}{\bc{X}}
\newcommand{\gpLabels}{\bc{Y}}
\newcommand{\gpMean}{\b{\mu}}
\newcommand{\gpVar}{\b{\sigma}}
\newcommand{\regStiff}{\b{\hat{K}}}
\newcommand{\joints}{\b{q}}
\newcommand{\inertiaMatrix}{\b{\Lambda}(\joints)}
\newcommand{\x}{\cartPose}
\newcommand{\xg}{\cartPose_g}
\newcommand{\stiffness}{\b{K}}
\newcommand{\damping}{\b{D}}
\newcommand{\externalForce}{\b{f}_{ext}}
\newcommand{\maxVel}{\b{v}_{max}}
\newcommand{\disp}{\Delta\x}
\newcommand{\maxDisp}{\disp_{max}}
\newcommand{\maxForce}{\b{F}_{max}}
\newcommand{\couplingForce}{\b{F}_{c}}

\newcommand{\threshold}{\gpVar_{tr}}

\begin{document}
\title{Interactive Imitation Learning \\ of Bimanual Movement Primitives}

\author{Giovanni Franzese, Leandro de Souza Rosa, Tim Verburg, Luka Peternel and Jens Kober}

\maketitle


\begin{abstract}
	Performing bimanual tasks with dual robotic setups can drastically increase the impact on industrial and daily life applications. However, performing a bimanual task brings many challenges, like synchronization and coordination of the single-arm policies. This article proposes the Safe, Interactive Movement Primitives Learning (SIMPLe) algorithm, to teach and correct single or dual arm impedance policies directly from human kinesthetic demonstrations. Moreover, it proposes a novel graph encoding of the policy based on Gaussian Process Regression (GPR) where the single-arm motion is guaranteed to converge close to the trajectory and then towards the demonstrated goal. Regulation of the robot stiffness according to the epistemic uncertainty of the policy allows for easily reshaping the motion with human feedback and/or adapting to external perturbations. We tested the SIMPLe algorithm on a real dual-arm setup where the teacher gave separate single-arm demonstrations and then successfully synchronized them only using kinesthetic feedback or where the original bimanual demonstration was locally reshaped to pick a box at a different height.     
\end{abstract}


\section{Introduction}

Modern society is faced with the lack of workforce in various repetitive jobs like re-shelving products in supermarkets or handling heavy luggage in airports. Robots appear to be the most promising solution to mitigate the negative effects of the declining workforce and perform these various complex tasks \cite{garabini2020wrapp}. To work in variable and unstructured environments, robots must be dexterous and intelligent to quickly learn the job while interacting safely with other robots, objects, and humans. However, traditional task-specific robot programming by experts fails to achieve such dexterity and intelligence due to the time-consuming process and poor adaptability of tailored solutions.

Recent advances in machine learning, namely in \gls{lfd}, have enabled robots to learn directly from (non-expert) human demonstrations without needing complex task-specific programming or long and dangerous exploration.
Branching from \gls{lfd}, \gls{iil} approaches \cite{celemin2022interactive} allow human teachers to provide interactive demonstrations and corrections to the robot, exploring the advantage that the latter is much more sample-efficient than the former, thus reducing the burden on the human teacher.
\gls{iil} methods cover many feedback modalities (e.g., correction, evaluative, and qualitative), can be used to learn different models (e.g., policies and objective functions), and leverage several function approximators (e.g., \glspl{nn}, \glspl{dmp}, \glspl{hmm}, and \glspl{gp}).

While tasks which require only one arm have been explored extensively in the literature, more complex tasks which require a bimanual setup have only recently been targeted.
Among such tasks, picking large objects in unstructured environments \cite{garabini2020wrapp}, assisting the elderly \cite{mukai2010development, zhu2023you}, surgery tasks \cite{berthet2016hubot} or complex assembly tasks \cite{zhang2017peg} are shown to require dexterous bimanual setups.
Factory assembly, logistics, and household applications of bimanual robots have been known for decades \cite{smith2012dual, yamada1995development}.

However, the increased number of \glspl{dof} (the curse of dimensionality) implies an increased teaching complexity and the necessity of skilled human teachers who knows how to interface with the bimanual robotic platform. 

\begin{figure}[t] 
	\centering
	\includegraphics[width=0.9\linewidth]{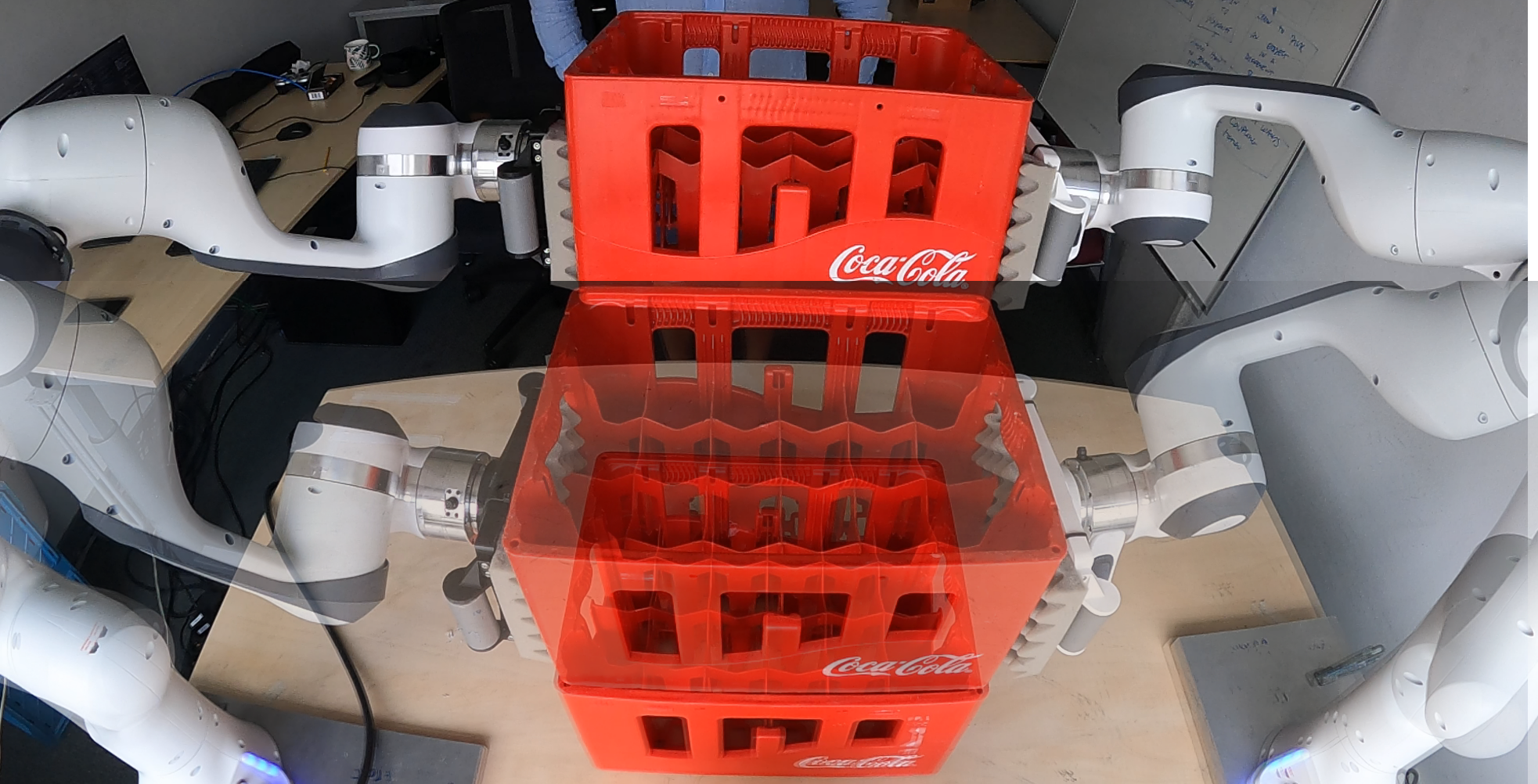}
	\caption{Example of possible application of bimanual manipulation: performing stacking of crates.
	}
	\label{fig:example_bimanual}
\end{figure}

In this paper we contribute with the \gls{simple} algorithm and propose:
\begin{itemize}
    \item The design of a bimanual impedance controller with variable Cartesian stiffness; safety constraints on the maximum applicable force and execution velocity are also formulated;
    \item A novel movement primitive formulation that allows efficiently learning long horizon tasks from a single demonstration and executes the motion in a reactive way; 
    \item Efficient corrections of the robot's policy directly from kinesthetic feedback, allowing for fine-tuning the demonstrations. Thanks to this, the user can show single arms' trajectories and fine-tune them when transferring the policies onto a bimanual task. 
\end{itemize}

To validate the proposed method, we conducted a series of experiments. The first three are technical experiments related to the main contributions that highlight and test different functionalities of the method. The last two are supplementary user studies to evaluate the type of data input for the proposed by comparing two human demonstration approaches and to evaluate giving corrections compared to giving new demonstrations. These additional insights can provide a better understanding of the input data generation method and adjustments of the robot's skill for bimanual cases.

\section{Related Works}\label{sec:related works}

\subsection{Bi-Manual Teaching Frameworks}

Like with single-arms, pre-planning and manual coding of multi-arm manipulation is a tedious process. An alternative is learning from \emph{human demonstrations}, where a user can guide the robot on how to execute the desired tasks.
However, when the user controls the dual (or multiple) robot setup, the physical and cognitive load increases drastically.
Using priors, shared control or task scaffolding, i.e., dividing the teaching into smaller parts, can substantially decrease the demonstrator workload and make the teaching easier and the learning faster.

Recent works on the control side of bimanual manipulation leverage \emph{shared control} strategies for reducing the burden of \emph{teleoperated} bimanual tasks.
For example, \cite{laghi2018shared} proposes a shared controller for helping the user to perform bimanual manipulation: it maintains the manipulators' relative position (or orientation) while the user controls the translations or rotations. Similarly, \cite{rakita2019shared} classifies human demonstrations in four teaching modalities: self hand-over, one-hand fixed, one-hand seeking, and fixed offset; when performing teleoperation, a trained classifier detects the most likely modality and adapts the constraints of the bimanual controller accordingly.

On the side of \emph{shared control}, \cite{tung2021learning} extends the Roboturk platform by having each arm teleoperated by a different teacher, reducing the cognitive load and enabling teaching tasks with more than two arms.
Moreover, ongoing research \cite{wen2022collaborative} presents a controller which enables inputs from a teleoperating user and local kinesthetic perturbations.
In our work, we focus on teaching bimanual policies from a single human teacher by teaching single-arm policies independently and then interactively reshaping them for successful coordination or adaptation to a new scenario. The goal is to enable non-expert users to teach complex bimanual tasks. 

\subsection{Bimanual Coordination Policies}\label{subsec: bimanual coordination policies}

During autonomous execution, disturbing one of the arms in a detached bimanual system can break the synchrony of the movements, making it necessary to provide both movement recovery and re-synchronization capabilities. 
The way the policy is encoded, e.g., time-dependent vs position-dependent, or the chosen function approximation, e.g., a \glspl{dmp}, \gls{hmm}, \gls{gp} \cite{williams2006gaussian} can change the disturbance rejection of the robot. 

To this end, the method in \cite{gribovskaya2008combining} uses a prior on the relative position of the two manipulators and a timing dependence in the \gls{hmm} formulation to synchronize the movement of arm manipulators.
Other approaches propose to create a ``leader and follower'' movement by adding a coupling term \cite{zhou2016coordinate}, a regulation term \cite{zhou2016learning}, or a deterministic encoding of trajectories with \glspl{dmp} \cite{saveriano2021dynamic}.
Alternatively, the epistemic uncertainty of \glspl{gp} can be used for switching the behaviour of the arms from follower to leader (and vice-versa) \cite{fanger2016gaussian}. This leader-follower learning paradigm makes the system react differently according to which arm is perturbed.
Alternatively, the task prior symmetry can be used for easily encoding and synchronizing the task. For example, \cite{bombile2022dual} proposes a bi-manual policy for picking and throwing non-stationary objects by learning a \emph{symmetric} dynamical system policy. In this case, perturbing any of the two arms would always make the other react.

Other approaches focus on achieving synchrony and coordination by segmenting the trajectories and reproducing them in sequence or according to a \emph{hierarchical} representation of the task.
The advantage of such approaches is that the sequencing provides an implicit synchronization on a higher level, making the lower-level problem easier. A common approach for this scheme is to learn policies for performing pre-defined sub-tasks, and a higher-level policy which creates a sequence from demonstrations \cite{kroemer2015towards, manschitz2015probabilistic}.
Alternatively, the task can have a pre-defined structure of sub-tasks based on heuristics, and synchrony is achieved with a sub-task scheduler \cite{caccavale2017imitation}.
Segmentation has also been used for deep-learning bimanual tasks in \cite{xie2020deep}, where lower-level policies are learned for each segment and higher ones for sequencing them. In this direction, \cite{mirrazavi2018unified} proposes a framework for multi-arm task-space control with smooth transitions from independent behaviors, e.g., when reaching goals, to dependent ones, e.g., when performing a dual-arm manipulation.

Our proposed approach differs from the approaches mentioned above in two ways.
First, these approaches fall under the \gls{lfd} category while our proposed \gls{simple} framework is an \gls{iil} algorithm, and to the best of our knowledge, \gls{simple} is the first framework for learning of bimanual tasks from interactive corrections.
Second, our interactive framework avoids heuristics for coordinating policies for each arm in a bimanual setup by using human feedback to regulate each arm's dynamics before transferring it to a bimanual policy.
Then, when the bimanual policy is executed, the robot's reaction to disturbances depends on the mechanical coupling of the end-effectors (see Section \ref{sec:cart imp cont}), or on chosen input state for the policy (see Section \ref{sec:model}).

\subsection{Motion Stability}

The stability of the bimanual operation is another key aspect. When learning from a small amount of data, in particular, the stability of the learned behaviour can be jeopardized when demonstrations are imperfect.
In \cite{gams2014coupling,likar2015adaptation}, a \gls{lfd} approach is combined with a learned controller that adapts the motion to keep the learned trajectory stable when facing external forces.
In \cite{bombile2022dual}, the motion is divided into one \gls{ds} for each sub-goal with a hand-designed vector field that brings the robot always close to the connecting lines of sub-goals.  
Our proposed \glspl{mp} have the objective of learning long-horizon \glspl{mp} with only one final goal and to obtain the stability property as an emerging behaviour of the motion encoding (Section \ref{sec:gp_stability}).

Next, Section \ref{sec:model} introduces the novel \gls{gp}-based formulation used for modeling \glspl{mp}, Section \ref{sec:SIMPLe} introduces the proposed \gls{simple} algorithm and how we use it for performing interactively learning bimanual \glspl{mp}, Section \ref{sec: robot validation} shows different applications and user-cases, and Section \ref{sec: conclusion} concludes the article with final remarks and future works.
\section{Movement Representation}\label{sec:model}

Section \ref{subsec: ggp} presents the proposed \gls{ggp} formulation,  Section \ref{subsec: ggp traj learning} the proposed trajectory learning framework and its benefits for safety, Section \ref{sec:gp_stability} presents the stability achieved with the proposed framework, and Sections  \ref{subsec: gp x ggp} and \ref{subsec: mov disambiguation} compare learning trajectories using traditional \glspl{gp} and the proposed \glspl{ggp}.

\subsection{Movement Learning with Gaussian Process}\label{subsec: ggp}

To learn the model of the demonstrated trajectories, we chose \glspl{gp} because it is a flexible non-parametric regression method where the kernel choice can be used to increase the inductive bias on the generalization of unseen points, what is prohibitive using function approximators such as \glspl{dmp} or \glspl{nn}.
Furthermore, its solid statistical formulation provides both the mean and the epistemic uncertainty of the prediction \cite{williams2006gaussian} what can be used for disturbance rejection or stiffness regulation \cite{franzese2021ilosa}.

Given the training data composed by a set of states $\gpStates$ and their respective labels $\gpLabels$, the prediction mean and variance at the evaluation point $\gpState$ follow, respectively:
\begin{equation}
	\label{eq::GP}
	\gpMean(\gpState) = \covPointTrain^\top\covTrainTrain^{-1} \gpLabels,
\end{equation}
\begin{equation}
	\gpVar(\gpState) = \covPointPoint -\covPointTrain^\top\covTrainTrain^{-1}\bc{k}_*,
\end{equation}
where the $\covPointPoint=\covPointPoint(\gpState,\gpState)$ is the variance of a single evaluation point $\gpState\notin\gpStates$, $\covPointTrain=\covPointPoint(\gpStates,\gpState)$ is the variance between $\gpState$ and the training inputs $\gpStates$, and $\covTrainTrain=\covTrainTrain(\gpStates,\gpStates)$ is the covariance matrix of the training data representing the leaned model \cite{williams2006gaussian}.  
Note that $\covPointPoint$, $\covPointTrain$, and $\covTrainTrain$ are based on the kernel functions and their hyper-parameters, which are used for incorporating prior knowledge into the process.

In particular, the kernel determines the interpolation and extrapolation behaviours and when using a distance-based kernel, e.g. \gls{rbf}, the prediction converges to the mean of the Gaussian Process, usually set to zero. 
Our objective is to have a mean function that extrapolates without losing the measure of epistemic uncertainty, i.e., does not return a vanishing prediction. 
For this reason, by correlating with only the closest neighbor, in the dataset and changing the kernel definition to: 
\begin{equation}\label{eq: kernel}
	\tilde{\covPointPoint}(\gpState_i,\gpState_j)= 
	\begin{cases}
		1 ,& \text{if } \covPointPoint(\gpState_i, \gpState_j) = \max(\covPointPoint(\gpState_i, \gpStates))\\
		0,              & \text{otherwise}
	\end{cases}
	\forall\gpState_j\in\gpStates.
\end{equation}

In simple terms, given a point $\gpState_i$, the correlation is 1 only if that is the maximum obtainable correlation when correlating $\gpState_i$ with all $\gpState_j \in \gpStates$. 
 With the new kernel, the prior covariance matrix becomes: 
\begin{equation}
\b{\Sigma}_{prior} =
    \left[ 
\begin{array}{c c} 
  \tilde{\covTrainTrain} &    \covPointTrain \\
   \graphCovPointTrain^\top & \covPointPoint
\end{array} 
\right].
\end{equation}

Note that, since the last column is for $\gpState_j\notin\gpStates$, the saturation is not applied. Thus, the resulting prior covariance matrix is not symmetric anymore, making the new process a pseudo-GP.
After the conditioning on the data points, the new pseudo-\gls{gp} posterior becomes: 
\begin{equation}\label{eq::GP_approx}
	\gpMean(\gpState) = \graphCovPointTrain^\top \tilde{\covTrainTrain}^{-1}\gpLabels = \graphCovPointTrain^\top\gpLabels,
\end{equation}
\begin{equation}
	\gpVar(\gpState) = \covPointPoint-\graphCovPointTrain^\top \bc{k}_\star.
\end{equation}

In simple terms, $\graphCovPointTrain$ selects as mean the label of the closest point in the database, computing the uncertainty according to the relative position between the query and the selected points.

Additionally, by saturating the covariance matrix $\covTrainTrain$, each trajectory element have its highest correlation with themselves: the new saturated correlation matrix, $\tilde{\covTrainTrain}$, is the identity matrix, thus eliminating the computationally heavy $\mathcal{O}\left(n^3\right)$ matrix inversion.
However, with this approximation, we are losing interpolation/smoothing properties. Meaning that the provided trajectory data must be without drastic jumps. In practice, recording trajectories with high enough frequency ($>10$Hz) and/or smoothing the data makes the use of the proposed approximation doable.
It is worthy to mention that the presented formulation is tailored for the specific application of movement learning and does not necessarily substitute general approximation methods like local models \cite{schneider2010robot} or variational approximations \cite{hensman2014scalable}. 
A detailed comparison between \glspl{gp} and \glspl{ggp} for trajectory learning is presented in Section \ref{subsec: gp x ggp}. 

\begin{figure}[t]
	\centering
	\includegraphics[width=0.9\linewidth]{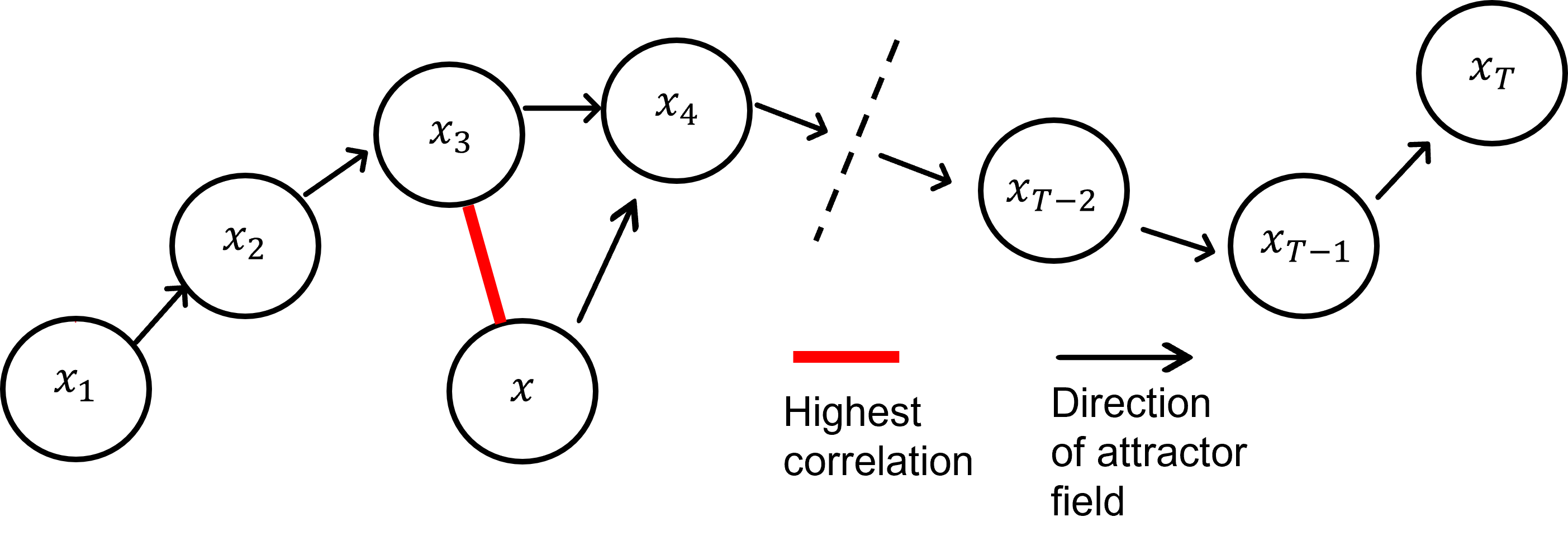}
	\caption{
		Representation of a trajectory as a chain of events. The state $\cartPose$ is the aggregation of the robot pose and time, where $\cartPose_i$ is $i$-th element in the reference trajectory. 
		Every element of the trajectory $i$-th has as element goal the next point on it $i+1$-th depicted by a forward arrow. The $\cartPose$ gets a unitary correlation with the closest element in the trajectory $m$-th and then as the goal the location of $(m+1)$-th state on the chain of events. The uncertainty is given by the distance from the  $\cartPose$ and its correlated point on the trajectory. 
	}
    \label{fig:graph_trajectory}
\end{figure}

\subsection{Representing Trajectories as Graphs}\label{subsec: ggp traj learning}

Our goal is to perform safe control during the general or corrective interactions between robots and humans.
To that goal, we start from a recorded trajectory demonstration, defined as an array of $n$ end-effector poses $\cartPoses = \{\cartPose_0, \hdots, \cartPose_{n-1}\} \in \mathbb{R}^3$ and the timestamp of each respective pose $\timestamps = \{\timeReal_0, \hdots, \timeReal_{n-1}\}\in\mathbb{R}$, and a final pose and time $\cartPose_n, \timeReal_{n}$, used to fit a policy $\policy$.
The trajectory can be seen as a sequence of events, represented as a graph with edges representing transitions from the state at time $ \timeReal_{i}$ to the state at $ \timeReal_{i+1}$.
Given the adopted \gls{gp} approximation, during the policy execution, the most correlated point is selected on the trajectory, and its label is selected as the goal, see Fig. \ref{fig:graph_trajectory}. We denote the policy as a \gls{ggp}.

However, the input type of the policy can completely change the robot behaviour. For example, a pose-only ``feedback" policy, $\pPolicy:\pPolicyState\rightarrow \pPolicyGoal$ is a fully reactive policy which computes the next Cartesian pose for the end-effector $\left(\pPolicyGoal\right)$, based on the current one $\left(\pPolicyState\right)$.
Such policies are safer since they make the robot to wait when its path is obstructed and allowing it to rejoin the trajectory on its closest point under perturbations \cite{franzese2021ilosa}. However, they cannot deal with movement ambiguities and time-dependent movements.

Alternatively, a time-only dependent policy, $\tPolicy:\tPolicyState\rightarrow \tPolicyGoal$, computes $\left(\tPolicyGoal\right)$ based on the current time $\left(\tPolicyState\right)$.
This type of policy can deal with movement ambiguities, e.g., when the demonstrated trajectory crosses itself, and with time-dependent movements, i.e., when the movement has to be temporarily paused at a specific position.
However, such ``feed-forward" policies are not a safe choice since the attractor moves on the trajectory without considering dangerous interactions with the human and with the environment. 

Instead, we proposed the usage of pose and time-\emph{belief} dependent policies, $\ptPolicy:\ptPolicyState\rightarrow \ptPolicyGoal$, which computes the pose goal and a new time belief $\left(\ptPolicyGoal\right)$ based on the current ones $\left(\ptPolicyState\right)$.
Note that the time-\emph{belief} is updated with the time of the selected goal in the trajectory. 
Encoding both pose and time belief allows for obtaining safe policies capable of handling time-dependent movements and ambiguities.

As such, \gls{simple} can be used with models fitted as time-dependent, pose-dependent, or pose and time-dependent policies by setting the \gls{ggp} states as $\gpState=\tPolicyState$, $\gpState=\pPolicyState$, or $\gpState=\left[\ptPolicyState\right]^\top$, respectively, and selecting a kernel for fitting the trajectories w.r.t. time $\left(\covPointPoint(\tPolicyState,\timestamps)\right)$, like in \cite{huang2019kernelized}, position $\left(\covPointPoint(\pPolicyState,\cartPoses)\right)$, like in \cite{franzese2021ilosa}, or both of them, as proposed in \gls{simple}, which is obtained by multiplying the time and the pose-dependent kernels, i.e., $\covPointPoint\left(\left[\ptPolicyState\right], \left[\cartPoses, \timestamps\right]\right) = \covPointPoint\left(\pPolicyState,\cartPoses\right) \circ \covPointPoint\left(\tPolicyState,\timestamps\right)$.

In the context of trajectory learning, the labels are set as the aggregation states in the demonstration which follow each state in the demonstration, i.e., $\gpLabels = \left[\cartPoses^d, \timestamps^d\right]^\top = \left[\{\cartPose_1, \hdots, \cartPose_{n}\}, \{\timeReal_1, \hdots, \timeReal_{n}\}\right]^\top$.

\subsection{Stability Analysis}
\label{sec:gp_stability}

From this \gls{ggp}-based formulation, we can also conclude that: 
\begin{proposition}\label{prop:stability}
	Using the trajectory graph representation, the motion always converge on the proximity of the demonstration and continues towards the end of it. 
\end{proposition}

\begin{proof}
	Since the vector $ \graphCovPointTrain^\top $ is correlating the current position of the end-effector with only one node of the trajectory, and if there is no overlap on the trajectory, the robot will move towards the goal of the closest node. Then, node by node, it  continues towards the end of the trajectory. 
\end{proof}

A great advantage of the pose and time trajectory encoding is that overlapping is no longer possible as the demonstrator cannot show two different robot positions simultaneously, leading to the absence of overlapping nodes, ambiguities, or undesired loops, guaranteing that the hypothesis in the proof of convergence is satisfied.
However, this also means that, when \emph{only} computing the correlation as a function of position, no physical overlapping of the trajectory can be demonstrated, such as when drawing an eight \cite{valletta2021imitation}.

\subsection{Comparison Between GPs and GGPs for Policy Learning}\label{subsec: gp x ggp}

\begin{figure}[t!]
	\centering
	\includegraphics[width=0.9\linewidth]{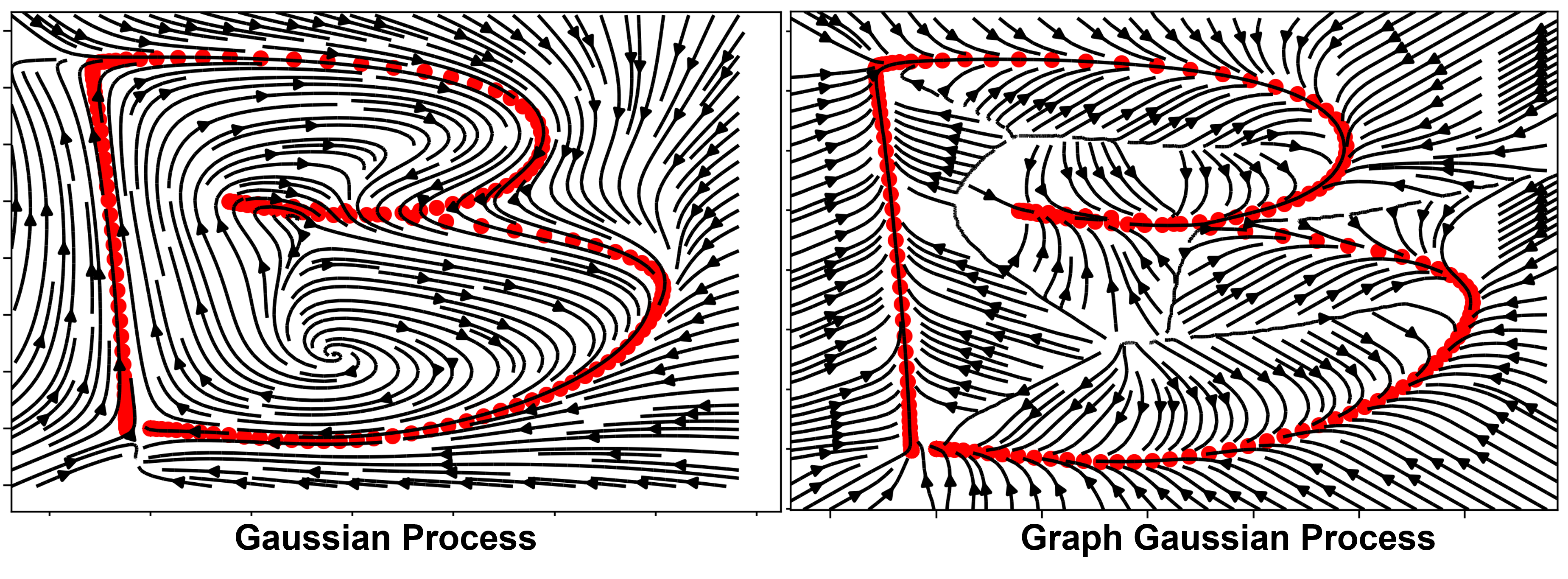}
	\caption{Comparison of the fitting of a trajectory with the shape of a ``B'' with \emph{position dependent} \gls{gp} and \gls{ggp}. The red dots are the recorded demonstrations, and the stream curves are the learned behaviour. }
	\label{fig:GP_vs_GGP}
\end{figure}

Figure \ref{fig:GP_vs_GGP} shows the different behavior in learning to draw the letter ``B'' (database from \cite{huang2019kernelized}) using a \gls{gp} and a \gls{ggp} using only the 2-D position.
The first thing to highlight is the effect of the kernel saturation in a faster convergence closer to the trajectory of the \gls{ggp} compared with the \gls{gp}. 
As consequence, when the robot is perturbed, the motion tends to go closer to the trajectory and continue from there. 
Nevertheless, this difference in the vector fields does not lead to unsafe sudden motions straight towards the attractor due to the proposed attractor and stiffness regularization/saturation described in Section \ref{sec:cart imp cont}.

The letter ``B'' shows a clear ambiguity at the overlapping of the trajectory between the two humps. The robot must first move in and then move out of the intersection on the same line in order to continue towards the end of the trajectory. The learned behavior of the two fitting methods is different. The \gls{gp} removes the overlapping ambiguity by considering it as noise. This results in cutting the motion without going down to the intersection of the curves, losing tracking accuracy.
On the other hand, in the line overlapping, the \gls{ggp} has a vector field pointing left when approaching from below and to the right when approaching from the top.
This may lead to an ambiguous situation that can cause the robot to get stuck locally or, in general, not track the motion correctly. This motivates the use of a position and time-dependent policy, to remove any possible state overlapping.

\subsection{Movement Disambiguation using Pose and Time-Dependent Policy}\label{subsec: mov disambiguation}

\begin{figure*}[ht]
	\centering
	\includegraphics[width=0.9\linewidth]{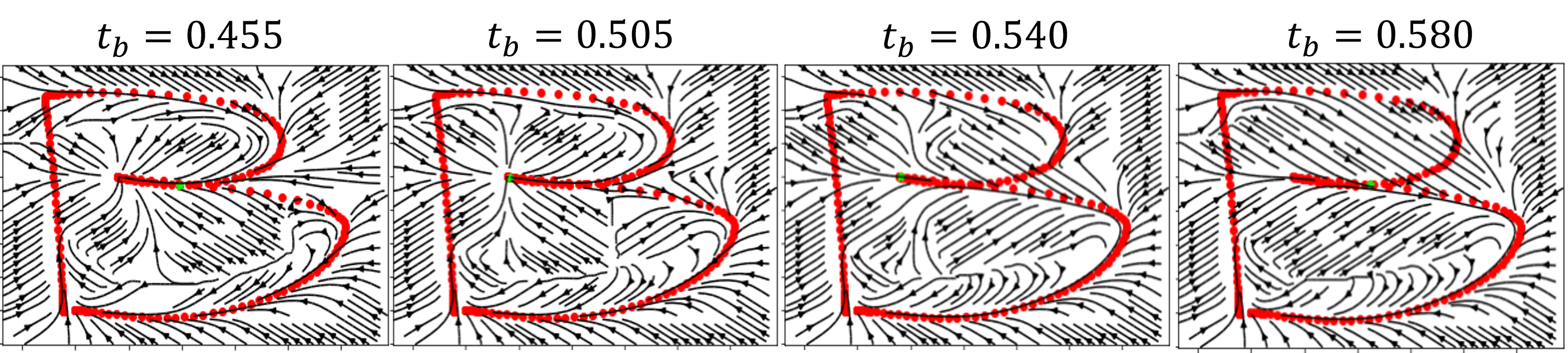}
	\caption{Attractor vector-field of a Graph Gaussian Process when the conditioning of the kernel with different time belief. $t_b$ is the time belief normalized by the total time of the trajectory, i.e., $0 \leq t_b \leq 1$. The corresponding element of the chain for every case is highlighted with a green circle. The red circles are the collection of points of the trajectory. In this example, the time belief is the time of the next element of the chain. }
	\label{fig:B_shape_time}
\end{figure*}

As explained in Proposition \ref{prop:stability}, no loops in the chain are allowed to guarantee good trajectory tracking. Thus, our solution is to consider also the time belief $\left(\timeBelief\right)$ in the state. Figure \ref{fig:B_shape_time} shows the evolution of the vector field for different time beliefs. The chain element of the trajectory for the $\timeBelief$ indicated above the figure is highlighted with a green dot. From the figure, it is possible to observe how the previously encountered ambiguity is elegantly solved. In fact, the robot gets into the valley and then out without getting stuck.

\begin{figure}[ht]
	\centering
	\includegraphics[width=0.9\linewidth]{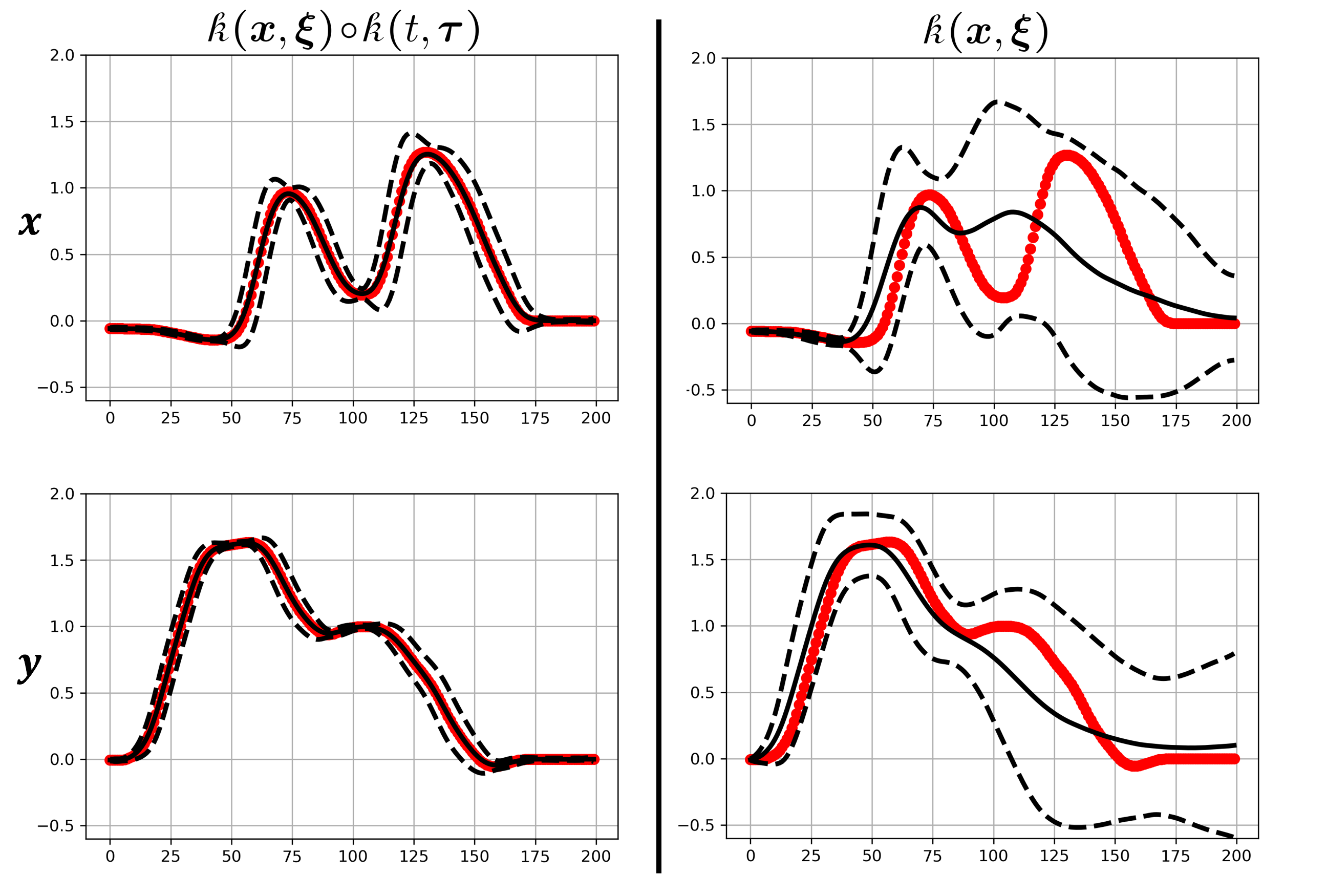}
	\caption{$200$ roll-outs of $200$ steps. The attractor position is injected with a Gaussian noise of zero mean and std of $0.01$. When the kernel is also taking into account the time belief, the motion is more robust when encountering ambiguity in the intersections of the two curves. Otherwise, ambiguity can lead to divergent behaviors. Red: original demonstration. Black: the average and standard deviation of the execution.}
	\label{fig:rollouts}
\end{figure}

In order to simulate the behaviors of a \gls{ggp} with or without a self-update of the time belief, $200$ different trajectories are rolled out starting from the origin of the demonstration. In order to take into account the inaccuracy of the low-level (impedance) controller, a Gaussian noise of magnitude $0.01$ is added to the attractor when computing the new position.
Figure \ref{fig:rollouts} depicts the mean and standard deviation of the trajectories. When the time dependence is active (left side of the figure), the trajectory always converges to the end, and the fluctuations are bounded. When only the position is considered (right side of the figure), the variability of the sampled trajectories increases, and the tracking is good on average until the start of the two humps intersection, from where the performance degrades due to the ambiguous states. 

\section{SIMPLe: Safe Interactive Movement Primitives Learning}
\label{sec:SIMPLe}

\begin{algorithm}[t!]
	\SetAlgoLined
	\DontPrintSemicolon
	\tcc{a) Passive Kinesthetic Teaching}

	\While{Trajectory Recording}{ \label{alg: demo begin}
		$\left[\cartPose,\timeReal\right]^\top =$ Receive$\left(\cartPose, \timeReal\right)$ \;
		
		$\gpStates = \gpStates \biguplus \left[\cartPose,\timeReal\right]^\top$ \;
	}\label{alg: demo end}
        $\quad \gpLabels_i = \gpStates_{i+1}, \forall i \in \{1, 2, \ldots, n-1\}$, with $n=\text{dim}(\gpStates)$
	
	\tcc{b) Active Kinesthetic Teaching}
	$\timeBelief=0$ \;
	\While{Control Active}{
		$\left[\cartPose,\timeReal\right]^\top =$ Receive$\left(\cartPose, \timeReal\right)$ \;
		$[\gpMean, \gpVar] =\ptPolicy(\cartPose, \timeBelief)$\;
		$[\ptPolicyGoal]^\top=\gpMean$\;
		$\timeBelief = \timeBelief_g$ \;
		Send($ \text{saturate}(\disp), \regStiff$)\tcp*{see Sec. \ref{sec:cart imp cont}}\label{alg: send 1}
		$\gpStates = \gpStates \biguplus \left[\cartPose, \timeReal\right]^\top$ \;
	}
        $\quad \gpLabels_i = \gpStates_{i+1}, \forall i \in \{1, 2, \ldots, n-1\}$, with $n=\text{dim}(\gpStates)$

	\tcc{c) Execution}
	$\timeBelief=0$ \\
	\While{Control Active}{
		Receive$\left(\cartPose\right)$ \;
		$[\gpMean, \gpVar] =\ptPolicy(\cartPose, \timeBelief)$\;
		$[\ptPolicyGoal]^\top=\gpMean$\;
		$\timeBelief$ = $\timeBelief_g$ \;
		Send($\text{saturate}(\disp), \regStiff$)\tcp*{see Sec. \ref{sec:cart imp cont}}\label{alg: send 2}
	}
	\caption{SIMPLe}
	\label{algo::interactive_teaching}
\end{algorithm}

The proposed \gls{simple} framework summarized in Algorithm \ref{algo::interactive_teaching} consists of three main parts. 
First, the human teacher provides kinesthetic demonstrations (Section \ref{sec:demonstrations}), from which a time and position-dependent model (Section \ref{sec:model}) is learned.
Second, the proposed method enables the human to provide demonstrations and to make interactive corrections (Section \ref{sec:demonstrations}), which are leveraged for learning the trajectories and synchronization of bimanual tasks (Section \ref{sec:bimanual learning}).
And third, the bimanual task can be executed. We employ a Cartesian impedance control to facilitate physical interactions during demonstrations, corrections and autonomous execution (Section \ref{sec:cart imp cont}), safety is ensured thanks to the proposed stiffness regulation \ref{subsec: stiff regulation} and coupling between manipulators \ref{subsec: dual cartesian impedance controller}.

Our method aims to enhance the teaching ability of non-expert users while guaranteeing a safe interaction while teaching, correcting, and executing bimanual tasks.
To cope with the complexity of teaching bimanual tasks, \gls{simple} provides an interactive \gls{kt} approach allowing to teach one arm at a time and then to teach how to synchronize them using touch by leveraging the time and pose-dependent \gls{ggp} formulation presented in Section \ref{sec:model}.
To the best of our knowledge, \gls{simple} is the first framework to employ \gls{iil} on bimanual setups.
Nevertheless, \gls{simple} does not restrict users from teaching (and correcting) both arms simultaneously, and it can be applied for single-arm manipulation tasks without any loss of generality.

\subsection{Teaching from Kinesthetic Demonstrations and Corrections}\label{sec:demonstrations}

\gls{lfd} allows non-expert users to program robots to perform complex tasks without any programming knowledge.
Different interfaces can be used to transfer data to the robot, such as teleoperation devices, touch screens or physical interaction with the robot's embodiment, obtaining a \gls{kt} approach.
When the user is teaching a task, the stiffness and damping of the Cartesian impedance controller are set to zero, allowing the user to easily move the robot.
The positions $\cartPoses$ and times $\timestamps$ of the demonstrated trajectories are recorded, and their respective goals, $\cartPoses^d$ and $\timestamps^d$ are obtained by shifting $\cartPoses$ and $\timestamps$ forward in time (Alg. \ref{algo::interactive_teaching}, lines \ref{alg: demo begin} to \ref{alg: demo end}).

After learning the motion from a kinesthetic demonstration, the user can reshape the trajectory of each arm to achieve, for example, coordination between the arms in the execution of the task. Given the Cartesian impedance controller (see Section \ref{sec:cart imp cont}) kinesthetic corrections can be performed by simply appling an external force .
Such a controller allows for the human to be in full control if the stiffness is set to zero, or the robot can gradually increase its control by regulating the stiffness. 

Additionally, given the time and pose-dependent policy (see Sec. \ref{sec:model}), the demonstrator can also drag the robot forward or backwards in time along its trajectory. 
This property can be used, for example, to make the execution of the initial demonstration faster \cite{kastritsi2018progressive, Meszaros2022RA-L}, to make the robot throw objects \cite{bombile2022dual}, or for synchronization learning, as proposed in this paper.

\subsection{Interactive Learning of Bimanual Tasks}
\label{sec:bimanual learning}

When teaching bimanual tasks, it is not always easy or feasible to provide kinesthetic demonstrations with both arms simultaneously, especially when using large redundant manipulators. Additionally, even when skilled users are able to teach a bimanual task by moving each end-effector with a single hand, they may perform a sub-optimal trajectory, or an ineffective one, given the task complexity.

In \gls{simple}, the movement of each arm can be executed independently according to the \gls{ggp} formulation described in Section \ref{sec:model}. 

The proposed interactive learning method offers many possibilities for non-expert users to teach complex bimanual tasks.
For example, they can demonstrate the movement for picking up a box one arm at a time and then learn to coordinate the two independent trajectories and apply enough pressure on the sides of the box to execute the task successfully.
Moreover, learning repetitive tasks like object hand-over can also be initially demonstrated one arm at a time and later use kinesthetic corrections to learn how to coordinate both arms.
Thanks to the calculation of the model as a function of position and time (belief), the user can also bring the robot back to the start of the trajectory and teach (with minimum interaction effort) to perform the task multiple times. 

\subsection{Safe Cartesian Impedance Control}\label{sec:cart imp cont}

\begin{figure}
	\centering
	\includegraphics[width=.7\linewidth]{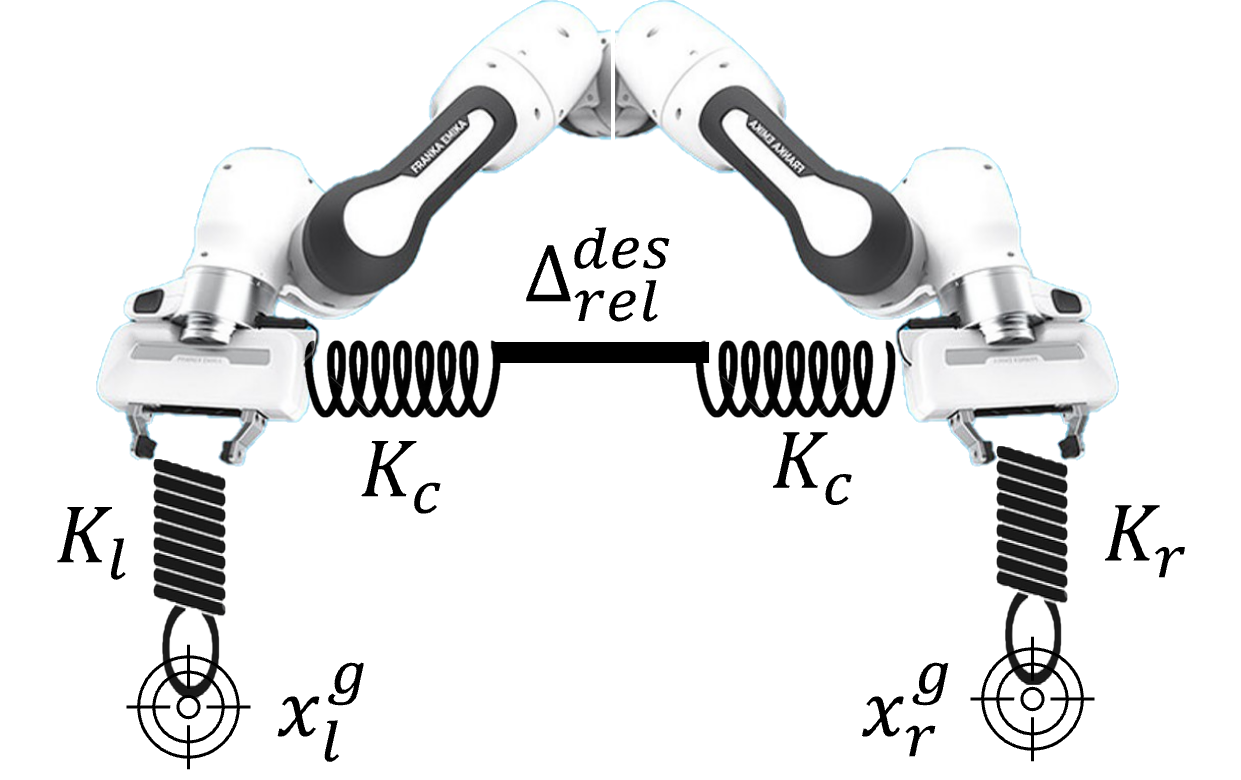}
	\caption{The Bimanual Impedance controller. For simplicity the spring-damper system is represented only with a spring. }
	\label{fig:bimanual_impedance}
\end{figure}

Figure \ref{fig:bimanual_impedance} illustrates the employed Cartesian impedance controller in each of the two manipulators, which emulates the behavior of a  mass-spring-damper system: 
\begin{equation}\label{eq:impedance}
	\inertiaMatrix\ddot{\x} = \stiffness\Delta \x-\damping\dot{\x} + \b{f}_{ext},    
\end{equation}
where $\inertiaMatrix$ is the Cartesian inertia matrix of the physical system,
$\stiffness$ and $\damping$  are
stiffness and damping that symmetric, positive-definite matrices; $\externalForce$ is the total external force, and $\Delta \x = \xg-\x$ is the distance between the goal and end-effector poses.
The damping matrix can be designed to simulate a critical damping system  \cite{coleman2023damping}.
In this framework, after computing the orthogonal decomposition of $\stiffness$, i.e., $\stiffness=\bm{R}\tilde{\stiffness}\bm{R}^T$ \footnote{$\bm{R}$ is an orthogonal matrix, hence $\bm{R}^T=\bm{R}^{-1}$}, where $\tilde{\stiffness}$ is a diagonal matrix, then $\tilde{\damping}=2\tilde{\stiffness}^{1/2}$ and $\damping=\bm{R}\tilde{\damping}\bm{R}^T$. Please notice that $\tilde{\stiffness}$ is a diagonal matrix, hence the square root is applied to every element of the diagonal.

The Cartesian impedance controller takes as input a stiffness matrix and a displacement vector (lines \ref{alg: send 1} and \ref{alg: send 2} in Algorithm \ref{algo::interactive_teaching});
in order to enhance safety  when interacting with humans \cite{haddadin2009requirements}, it is necessary to saturate the attractor displacement and the stiffness to a maximum safe value. To help the bounds definition, we can compute them as a function of the desired maximum free-movement velocity $\left(\maxVel\right)$ and maximum applicable static force of the end-effector$\left(\maxForce\right)$ (in absolute values). 

First, we compute an upper bound for the maximum displacement.
Considering Equation \eqref{eq:impedance}, when the robot is in free-movement, i.e., $\externalForce=0$, the maximum velocity happens for $\ddot{\b{x}} = 0$, that is to say:
\begin{equation}\label{eq: max vel}
	{\damping} |\dot{\x}| = {\stiffness}  |\disp|.
\end{equation}

Thus, given the current setted stiffness $\stiffness$ and the desired max allowed velocity $\maxVel$, $\Delta \x$ needs to respect:
\begin{equation}\label{eq:vel_saturation}
    |\disp| \leq \maxDisp =\stiffness^{-1} \damping \maxVel =2 \bm{R} \tilde{\stiffness}^{-\frac{1}{2}} \bm{R}^T \maxVel,
\end{equation}
obtained after using the definition of damping. Before sending to the robot, the $\disp$ is \emph{saturated} in order to respect the upper bound. 

However, if taking into account the maximum static force ($\maxForce$) when $\dot{\x}=0$ and $\ddot{\x}=0$, an upper bound on the stiffness can be found, such that:
\begin{equation}
	\stiffness \maxDisp \leq \maxForce;
\end{equation}
\begin{equation}
	\bm{R}\tilde{\stiffness}\bm{R}^T\maxDisp \leq \maxForce;
\end{equation}
\begin{equation}
	2 \tilde{\stiffness}\bm{R}^T \bm{R} \tilde{\stiffness}^{-1/2} \bm{R}^T \maxVel  \leq \bm{R}^T\maxForce;
\end{equation}
\begin{equation}
     2 \tilde{\stiffness}^{\frac{1}{2}} \b{R}^T \maxVel  \leq \b{R}^T\maxForce.
\end{equation}

Hence, since the matrix $\tilde{\stiffness}$ is diagonal, we can find the upper bound of each element in the $i$-th row and column $\left(\tilde{\stiffness}_{ii} \right)$ as:
\begin{equation}
\label{eq:stiff_saturation}
     \tilde{\stiffness}_{ii} \leq \left(\frac{(\b{R}^T\maxForce)_i}{2 (\b{R}^T\maxVel)_i}\right)^2.
\end{equation}
so, in every singular component, the value of the principal stiffness is \emph{saturated} in order to respect the found inequality. 

\subsection{Stiffness Regulation}\label{subsec: stiff regulation}

Regulating the stiffness can be used to incrementally increase the stiffness after each demonstration, reducing human control as the learned movement is interactively refined \cite{tykal2016incrementally}. Alternatively, the stiffness can be regulated when perceiving strong external forces, as a disagreement detection \cite{van2022disagreement}. Similarly, \cite{kastritsi2018progressive} proposed a variation of a \gls{dmp} where the robot variable stiffness and the regressor phase are modulated to adapt to human kinesthetic demonstrations. 

When more demonstrations are provided, the measure of \emph{aleatoric} uncertainty, i.e., variability in the demonstration, can be used to regulate the tracking stiffness of the robot \cite{jaquier2020learning}.
Differently, we propose to exploit the \emph{epistemic} uncertainty quantification of the policy $\left(\gpVar\right)$, enabling for automatically regulating the Cartesian impedance controller's stiffness, hence switching control between robot and human.

Mathematically, 
\begin{equation}\label{eq:regulation}
	\regStiff = \text{saturate}(\stiffness) \frac{1-\gpVar(\x)}{1-\threshold},~\text{when~} \gpVar(\x) > \threshold 
\end{equation}

where the $\threshold$ is the uncertainty threshold that is used to detect the disagreement. Note that $\gpVar(\x)$ goes from $0$ when close to the trajectory, to $1$ when at infinite distance from it.
Thanks to this stiffness regulation, when the robot is dragged in regions of high uncertainties, it mitigates the external force applied to the user perturbing the trajectory.
This behaviour can be conceptualized as the robot's non-verbal teaching request or repositioning into regions closer to the demonstration. 

\subsection{Dual Cartesian Impedance Control}\label{subsec: dual cartesian impedance controller}

Differently from the execution of a single-arm, when a two-arms policy execution is performed, extra attention is required regarding the
mechanical coupling of the movement. For example, when picking up a box with two hands and executing a re-shelving operation, in case of a perturbation of one arm, the other arm must also follow the perturbed movement.
In this case, both arms must be \emph{mechanically} coupled, meaning that in the impedance control of each arm, we would add an extra coupling force defined as:  
\begin{equation}\label{eq:coupling_left}   
	\couplingForce^l = \stiffness_{c}\left(\x_r-\x_l-\disp_{rel}^{des}\right)+\damping_{c} \left(\dot{\x}_r-\dot{\x}_l\right),
\end{equation}
\begin{equation}\label{eq:coupling_right}
	\couplingForce^r = \stiffness_{c}\left(\x_l-\x_r+\disp_{rel}^{des}\right)+\damping_{c} \left(\dot{\x}_l-\dot{\x}_r\right),
\end{equation}
where $\Delta \x^{des}_{rel}=\x_r^{des} - \x_l^{des}$ is the desired distance from the two end-effectors controlled by the \gls{simple} algorithm. A simple schematic visualization of the proposed bimanual impedance control is displayed in Fig. \ref{fig:bimanual_impedance} where each end-effector is coupled with a stiffness (and damper) with respect to their goal but also with a relative stiffness (and damper) between them.

Note that the proposed safety saturation and regulation process described, respectively, in Section \ref{sec:cart imp cont} and Equation \eqref{eq:regulation} are applied on a per-arm basis, thus being applied to single-arm setups.
For a bimanual setup, the displacement and stiffness for the coupling forces ($\couplingForce$, defined in Equations \eqref{eq:coupling_left} and \eqref{eq:coupling_right}) are saturated and regulated similarly to Equations \eqref{eq:vel_saturation}, \eqref{eq:stiff_saturation}, and \eqref{eq:regulation}.

\section{Real Robot Validation}\label{sec: robot validation}

We performed the experiments with two 7-\gls{dof} Franka-Emika Panda placed vertically on a table and with the same orientation. 
The impedance control was implemented\footnote{\url{https://github.com/franzesegiovanni/franka_bimanual_controllers}} as described in Section \ref{sec:SIMPLe}. Each manipulator had a shared memory of their Cartesian poses, allowing the calculation of the mechanical coupling force. The experiments presented in Sections \ref{subsec:validation teleop x kines}, \ref{subsec:validation asynch crate}, and \ref{subsec:validation synch crate} were performed using a custom 3D-printed plate end-effector depicted in Fig.~\ref{fig:user_study}, which features a layer of soft form for reducing the interaction forces during impacts with objects as in \cite{dehio2022enabling}; the experiment presented in Section \ref{subsec:validation handover} was performed using the Franka gripper. The impedance control framework, written in C++ makes use of Robot Operating System (ROS) to interface with Alg. \ref{algo::interactive_teaching}, written in Python.

We perform $5$ experiments with the real robot setup:
\begin{enumerate*}[label=\roman*)]
	\item The interactive synchronization of the picking motion of a bottle crate when the demonstration is provided separately for each robot, showing how \gls{simple} is used to learn a bimanual synchronization, 
	\item the interactive correction in picking a different crate compared to the one of the original demonstration, showing how to use the \gls{ggp} formulation to modify the motion locally, 
	\item a handover task, where one robot picks and places an object and the other robot picks it from the other's goal location and places it at another position, showing the ability to restart the execution of a trajectory simply dragging the robot at the starting location,
    \item a supplementary user study to compare teleoperation and \gls{kt}, the two most common types of demonstration approaches,
    \item a supplementary user study to compare giving interactive corrections to giving new demonstrations.
\end{enumerate*}

The first three are technical experiments to highlight and validate different functionalities of the proposed method.
\begin{scontents}[store-env=description_experiments] 
\high{
Each experiment was conducted in $5$ trials, and for each of them, the final learned motion was performed $5$ times after demonstration and correction(s). This approach allowed for the assessment of the reliability of the learned skill.
}\end{scontents}
The last two are supplementary user studies to evaluate the type of data input for the proposed by comparing two human demonstration approaches and to evaluate giving corrections compared to giving new demonstrations. These additional insights can provide a better understanding of the input data generation method and adjustments of the robot's skill for bimanual cases.
For all the experiments, we used a position-time kernel for the \gls{ggp} that computes the correlations and updates the time beliefs online.
We use a negative exponential kernel, i.e. $k= \exp\left(-\frac{|x_i-x_j|}{\lambda}\right)$, with a length scale of $0.05~\mathrm{m}$  for the space correlation and $0.05~\mathrm{s}$ for the time correlation. The sigma threshold is set to $\sigma(\lambda)$ which is the uncertainty when the closest point is at distance $\lambda$. The Cartesian stiffness is kept to $600~\mathrm{N/m}$ for linear stiffness and $30~\mathrm{Nm/rad}$ for rotational. The attractor distance is saturated at $0.05~\mathrm{m}$, implying that the expected maximum applicable force is $30~N$ in every linear Cartesian direction and the maximum expected linear velocity is $\approx 0.6~\mathrm{m/s}$ in every linear direction. The rotation delta is saturated at $0.15~\mathrm{rad}$, implying a maximum torque of $4.5~\mathrm{Nm}$ in every rotational component and a maximum velocity of $\approx 0.4~\mathrm{rad/s}$. The coupling stiffness is set to $800~\mathrm{N/m}$ in the linear components and $0$ for the rotational ones. The relative error is also saturated at $0.05~\mathrm{m}$.
A video of the experiments can be found at: \\ \texttt{\url{ https://youtu.be/GasxgbJZHdQ }}.

\subsection{Asynchronous Crate Picking}\label{subsec:validation asynch crate}

\begin{figure*}
	\centering
	\includegraphics[width=0.9\linewidth]{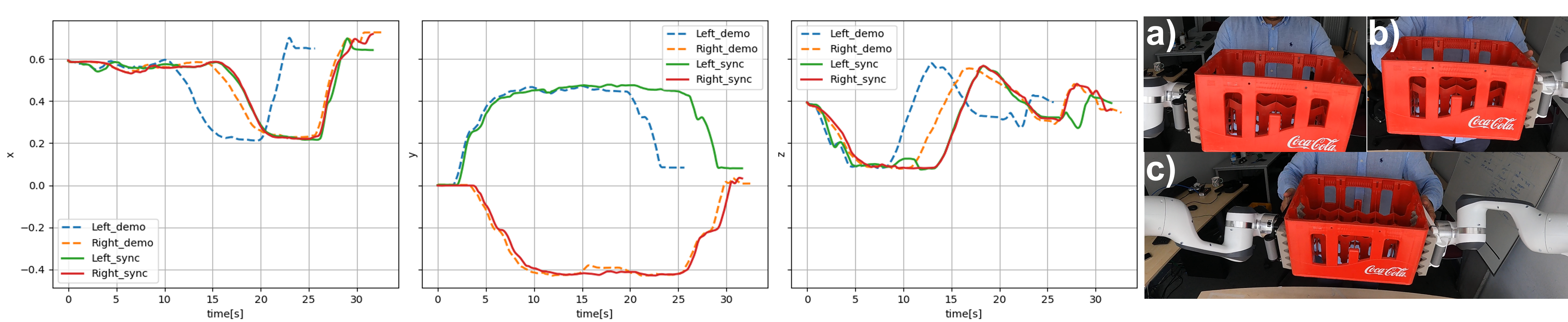}
	\caption{Interactive synchronization of a bimanual picking task. The dashed lines are the demonstrations recorded in the independent demonstration phase a) and b). Since they are not perfectly synchronized, the autonomous execution would fail, hence, the human feedback in c) allows a successful synchronization, depicted with solid lines.}
	\label{fig:syncronization_robot}
\end{figure*}

When a pianist approaches studying a new piece, they do it \emph{one hand at a time}. After mastering the movement with each hand, they start learning how to successfully coordinate the combined execution. 
Inspired by this idea, in this validation experiment, the user is asked to demonstrate how to best pick a crate, first with the right and then with the left manipulator. However, when the independently learned behaviours were executed with \gls{simple} the coordination was off, and the handling of the crate was not stable. In Section \ref{sec:bimanual learning}, we highlighted how user feedback can be used to reshape the trajectory and that the reactive formulation of \gls{simple} makes the trajectory to ``virtually'' stop: this feature can be used to learn a bimanual task while simply coordinating the separately recorded policies, see Figure \ref{fig:syncronization_robot}.

The effect of the human input can be appreciated in Figure \ref{fig:syncronization_robot}. The original demonstrations are represented by dashed lines. Even if the movement of the two demonstrations looks correctly symmetric with respect to the y-plane, the right arm is slower. However, it can be noticed how, after only one correction round, the motion of the two demonstrations is synchronized, as depicted with a solid line. Given the perfect obtained synchronization, in the next round, the user focused on increasing the applied pressure on the side of the crate to increase the grasp reliability.  
\begin{scontents}[store-env=experiment_1] 
\high{In the 5 experiment repetitions, the user consistently provided necessary synchronization corrections. One trial had an additional correction round, and two trials had two extra correction rounds. After the interactive correction rounds, the robot always placed the crate correctly. The Cartesian error of the final crate position with respect to the final round of correction, considering 25 repetitions (5 executions x 5 trials), has a mean of $0.021~\mathrm{m}$ and a standard deviation of $0.009~\mathrm{m}$.}
\end{scontents}

\subsection{Synchronous Crate Picking}\label{subsec:validation synch crate}

\begin{figure*}
	\centering
	\includegraphics[width=0.9\linewidth]{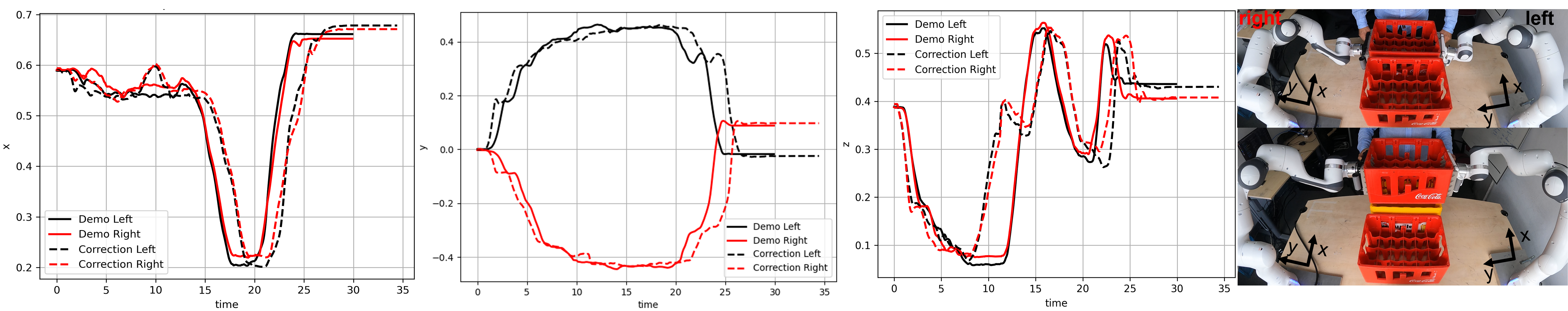}
	\caption{ Use interactive learning to teach the robot how to modify the original trajectory so the robot can learn how to pick a crate that is at a different height. }
	\label{fig:correction_box}
\end{figure*}

In this experiment, we focused on successfully teaching the same task of picking a box but giving bimanual demonstrations and corrections. In particular, we showed that even giving only one bimanual demonstration with a few rounds of corrections, the task execution was successful. We also tested the possibility of locally modifying the original policy to pick a different box placed at a higher level. Figure \ref{fig:correction_box} highlights how the robot can be dragged higher sooner, at around 10 seconds, and how, after picking the crate, the robots follow the original policy, being able to place the crate and go to resting position autonomously. 
\begin{scontents}[store-env=experiment_2] 
 \high{In the 5 experiment repetitions, in the first two trials, the user provided two rounds of correction, but only one in the last three. 
 The final position error of the box has a mean of $0.005~\mathrm{m}$ and a standard deviation of $0.004~\mathrm{m}$.}
\end{scontents}
It is important to notice that even knowing the box's position, the motion's generalization in a task-parameterized approach is not trivial. In fact, the policy would have to move with respect to the picking frame and then, after a successful pick, switch with respect to the goal crate. This logic has shown to be successfully implemented in \cite{Meszaros2022RA-L} but also to be a source of generalization ambiguities \cite{franzese2020learning}.
In general, performing a shared controlled teaching, with the user only taking control locally, can drastically reduce the burden of giving new complete demonstrations. 

\subsection{Object Hand-over}\label{subsec:validation handover}

Another example of a tedious task is repetitive demonstrations: being able to demonstrate the task only once and then interactively assemble a long trajectory allows the teaching of complex bimanual coordination tasks, like stirring a coffee mug \cite{sun2022mixline} or learning a handover task. To validate \gls{simple} in this circumstance, we taught the right arm to pick up a box and place it on the central separation line between the two robots. Then, the left arm would pick up the box and place it in its front. 
The goal is to show how dragging the robot around can be used for re-synchronization or local trajectory reshaping and also as a movement ``reset''. 

\begin{figure*}
	\centering
	\includegraphics[width=0.9\linewidth]{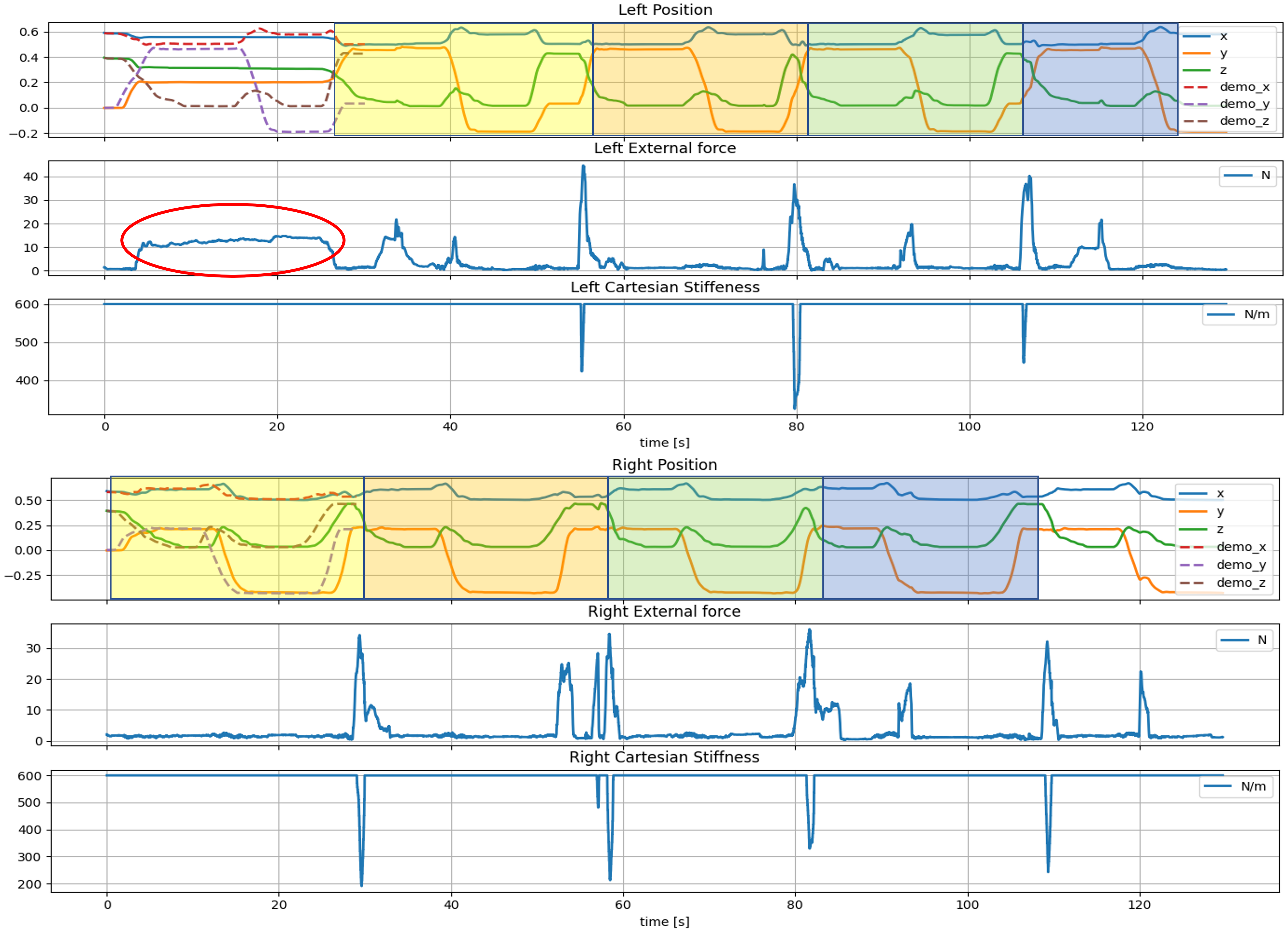}
	\caption{Object hand-over. For both the left and the right arm, the Cartesian position (x,y,z) is depicted as a function of time, the Cartesian linear stiffness, and the norm of the Cartesian external force. The red circle indicates the region where an external force was applied to the left manipulator to correct the trajectories.}
	\label{fig:hand_over}
\end{figure*}

The original demonstrations are displayed with a dashed line in Figure \ref{fig:hand_over}.
When executing the motion with \gls{simple}, the human can safely apply a force on the robot to stop its execution or drag it around on another desired position of the motion. 
At the beginning of Figure \ref{fig:hand_over}, a force is applied to the left manipulator (highlighted by a red circle) to temporally stop it from moving, allowing the right arm to successfully pick a box and place it on the center line. At the moment that the user releases the robot, it is free to move and can pick up the box and reach its goal. To allow the repetition of the motion, the user applies a larger external force (observable with peaks), causing a drop in stiffness since the robot is probably dragged into a region of space with a lower correlation according to \eqref{eq:regulation}. Every time the robot finishes its pick and place task, if the user is willing to repeat it, they only have to drag the robot to the desired position of the trajectory. The user is teaching the motion multiple times, as reported with colored patches in the figure.
\begin{scontents}[store-env=experiment_3]
\high{
We measure the final error in placing the box after the handover, executed 5 times in 5 different demonstration trials. The mean error and standard deviation are $0.011~\mathrm{m}$ and $0.008~\mathrm{m}$, respectively. }
\end{scontents}

\subsection{User study: Teleoperation vs. Kinesthetic Teaching}\label{subsec:validation teleop x kines}

The algorithm itself works with different data from different types of demonstrations. However, since obtained input data depends on the type of demonstration, the demonstration method is an essential part of the whole framework. Therefore, we conduct a supplementary user study to provide additional insight into the effects of the demonstration method to compare the two most common demonstration approaches: teleoperation and kinesthetic guidance. There are studies comparing both teaching approaches, but they were conducted for a single arm \cite{fischer2016comparison,gliesche2020kinesthetic}. The study in this paper looks into this subject from a bimanual perspective.

Section \ref{sec:related works} highlighted how different works focus on enhancing the teleoperation ability of non-expert users using assistive techniques like shared autonomy \cite{laghi2018shared, rakita2019shared, tung2021learning}. 
Since \gls{simple} works with both teleoperated and kinesthetic demonstrations/corrections, we wanted to study which is more user-friendly. Although, getting the true answer is not easy: the teleoperation device can have a strong influence, as well as the dimension of the robot or the requested task. 
For the conducted user study, we asked 7 non-expert users to perform a relatively simple task: pick a box and stack it on top of another. These 7 users were all male and with ages ranging 23 and 40 years old. 
In order to mitigate the learning bias from the results, participants had a familiarization phase for each teaching modality, in which they could restart the teaching session up to 5 times. 
For every new participant, the first teaching modality was alternated between teleoperated and kinesthetic, to remove the bias due to their familiarization with the task.

For metrics, we measured the success rate in solving the task and the total teaching time for each method. For subjective analysis, we asked the participants to complete a NASA TLX questionnaire.
We conducted a paired samples t-test to verify if the time to do \gls{kt} is significantly shorter than for teleoperation with the 6D mice. However, 3 people out of 7 failed to perform successful teleoperation, because they did not manage to coordinate well, making the robot self-collide or reach a joint limit. Therefore, we set as failure time the maximum time of the non-failing ones.
The test showed that \gls{kt} requires less time compared to the teleoperation with the given hardware with the difference being statistically significant ($p < 0.05$). 

\begin{figure}
	\centering
	\includegraphics[width=\linewidth]{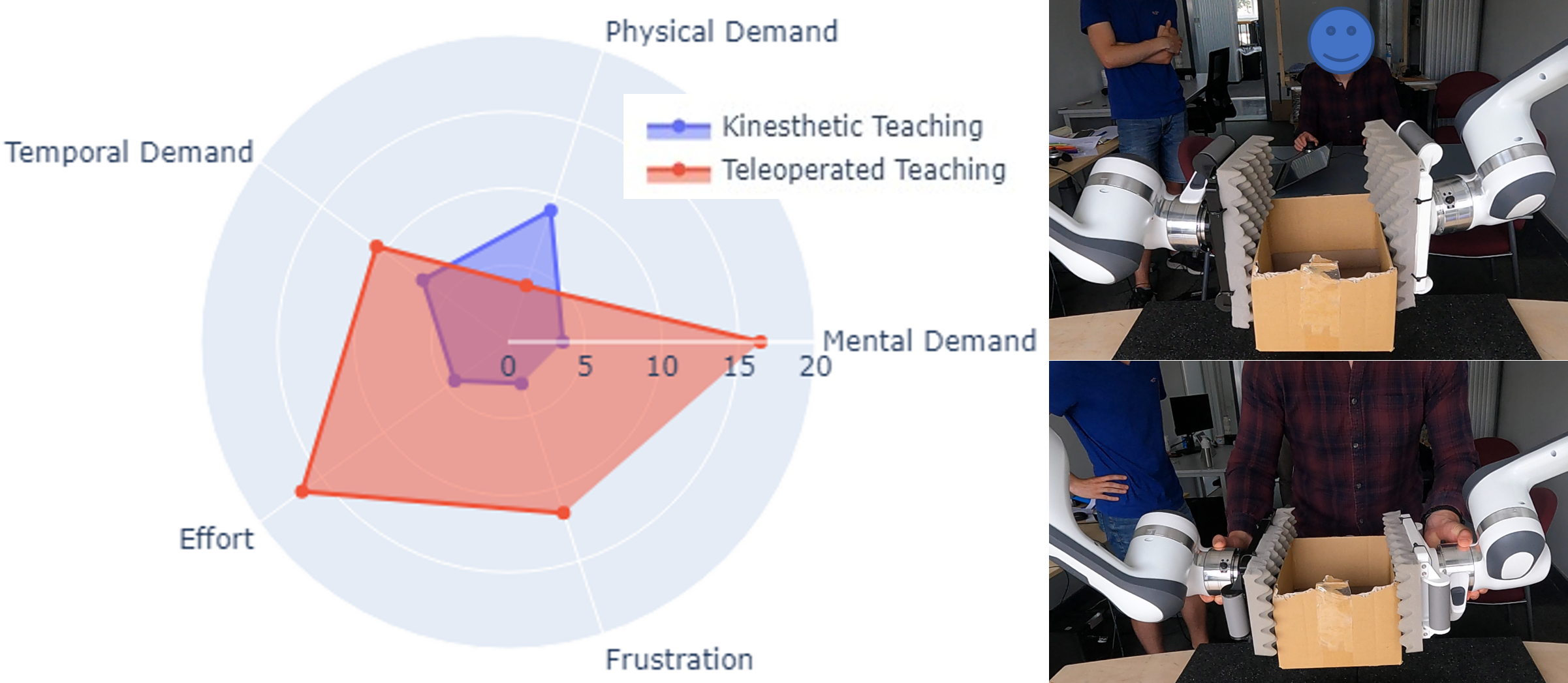}
	\caption{User study to compare the performance of non-expert users in performing bimanual teleoperation with two 6-D mice versus bimanual \gls{kt}. On the left are scores of the NASA-TLX questionnaire, and on the right are the set-ups of teleoperating and performing \gls{kt}.}
	\label{fig:user_study}
\end{figure}

Figure \ref{fig:user_study} illustrates the average NASA TLX scores among the different users. We can observe that teleoperation resulted in being more mentally demanding and frustrating to perform.
In general, we could observe that users tend to focus on teleoperating one arm at a time, making handling the box impossible. When providing \gls{kt}, the physical contact with the robot helps them to understand the best trajectory better and to accomplish the task successfully.  

\subsection{User study: Corrections vs. New Demonstration}\label{subsec:validation corr x new}

Besides the input data generation method, another key factor related to bimanual manipulation teaching is how humans correct existing skills and what is their preference between correcting or giving a new demonstration. To test this, 12 non-expert users participated in an experiment structured as follows: The user was asked to demonstrate the task of placing a box on the crate. The demonstration was then shown to the user after an offset was applied to the initial position of the box. The user was then tasked with kinesthetically correcting the initial policy to account for the change in the initial position. This was repeated two times for different initial positions of the box. The user now should have a sufficient understanding of what it means to give a demonstration or correction.

The second part of the experiment was designed to find the user preference for increasing lengths of the demonstration. The user was tasked with first demonstrating the task of placing the box on the crate. After the demonstration, an offset was applied to the box and the user was given the choice to either correct or re-demonstrate given the new initial condition. For the second iteration, the task remained the same with the additional requirement that after picking up the box, before placing it on the crate, the user has to move the box through a different location as a waypoint. This was done to artificially lengthen the demonstration. Once again an offset was applied to the initial position of the box and the user was given the choice between correcting or redemonstrating. This was done one last time with two waypoints. 

Given the choice, out of the 12 participants, 11, 8, and 10 chose to adjust the policy with the interactive corrections for the experiment with zero, one, and two waypoints, respectively, rather than providing new demonstrations. Thus, only in 7 out of the 36 trials, a new demonstration was preferred, which indicates a strong preference for interactive corrections. Afterwards, to evaluate their experience they were asked to answer several Likert scale questions related to user perception of corrected skill and their physical/mental load. The results can be seen in Table \ref{tab:likertRes}, where the number in each cell represents the number of participants that choose a particular agreement on the Likert scale.

\begin{table}[ht!]
\caption{Likert scale: corrections vs. new demonstrations}\label{tab:likertRes}
\begin{center}
\begin{tabular}{l|rrrrr}
Score & \multicolumn{1}{l}{Q1} & \multicolumn{1}{l}{Q2} & \multicolumn{1}{l}{Q3} & \multicolumn{1}{l}{Q4} & \multicolumn{1}{l}{Q5} \\ \hline
0 (strongly disagree)     & 0                      & 0                      & 0                      & 4                      & 0                      \\
1 (disagree)              & 0                      & 2                      & 2                      & 4                      & 2                      \\
2 (slightly disagree)     & 0                      & 2                      & 4                      & 3                      & 4                      \\
3 (slightly agree)        & 1                     & 3                      & 5                     & 0                      & 1                      \\
4 (agree)                 & 2                      & 4                      & 0                      & 1                      & 5                      \\
5 (strongly agree)        & 9                     & 1                      & 1                      & 0                      & 0                      \\ \hline
mean                      & 4,67             & 3.00                   & 2,50                   & 1,67                  & 2,75                   \\
standard deviation        & 0,65                   & 1,28                   & 1,09                  & 1,19                  & 1,22                 
\end{tabular}
\end{center}
\raggedright\footnotesize{ \vspace{2pt} Q1: After giving Kinesthetic demonstration, I feel the robot is performing the task well, \\Q2: After providing corrections, I feel the robot is adapting well to the novel situation, \\Q3: I feel that giving a bimanual Kinesthetic Demonstration with two arms is tedious for a human teacher, \\Q4: Performing Interactive Corrections is LESS physically tiring than giving a completely new demonstration,\\Q5: Performing Interactive Corrections is LESS mentally tiring than giving a completely new demonstration}
\end{table}

The users found that both new demonstrations and corrections were effective at improving the robot's task. The users were split on whether the bimanual demonstrations are tedious. In general, they found interactive corrections more physically demanding than providing new demonstrations, probably because the robots were already performing movements rather than being completely compliant during new demonstrations. During the experiments, it was observed that people that were shorter, had smaller hands, or were less muscular, tended to struggle more with correcting a policy. Those participants thus might have preferred giving a new demonstration over a correction. However, the users perceived interactive corrections as slightly less mentally demanding, probably because they needed to pay attention only to specific segments as opposed to the whole task.
\section{Conclusion}\label{sec: conclusion}

This paper contributes to the field of bimanual manipulation with an interactive kinesthetic learning framework named \gls{simple}. It uses a novel formulation of \gls{gp}, named \gls{ggp}, that is computationally efficient and ensures local and global stability of the motion while retaining an estimation of epistemic uncertainties. Thanks to the kernel formulation, the policy encoding can go from purely time-dependent to purely position-dependent or to a combination of both. At the same time, the graph representation of it allows an online update of the time \emph{belief} that, differently from the robot position, cannot be directly measured. The study reports a comparison of a \gls{gp} with the novel \gls{ggp}, see Figure \ref{fig:GP_vs_GGP} and an ablation study when the time dependence is considered or not, see Figure \ref{fig:rollouts}. We conclude that considering the time and properly updating its beliefs allows dealing with more complex and possibly ambiguous demonstrations. 

Various technical validation experiments were performed on a real bimanual setup to demonstrate the key functionalities and capabilities of the proposed method. The supplementary user studies gave interesting insights into how humans feel when teaching and correcting a robot with different modalities. Our study reported that users are faster and less stressed when performing kinesthetic teaching compared to teleoperation. Furthermore, most users prefer giving corrections to completely new demonstrations.

\section*{Acknowledgements}
This work was supported by the European Research Council Starting Grant TERI Teaching Robots Interactively, under Project 804907.

\bibliographystyle{IEEEtran}
\bibliography{paper}

\begin{IEEEbiography}
	[{\includegraphics[width=1in,height=1.25in,clip,keepaspectratio]{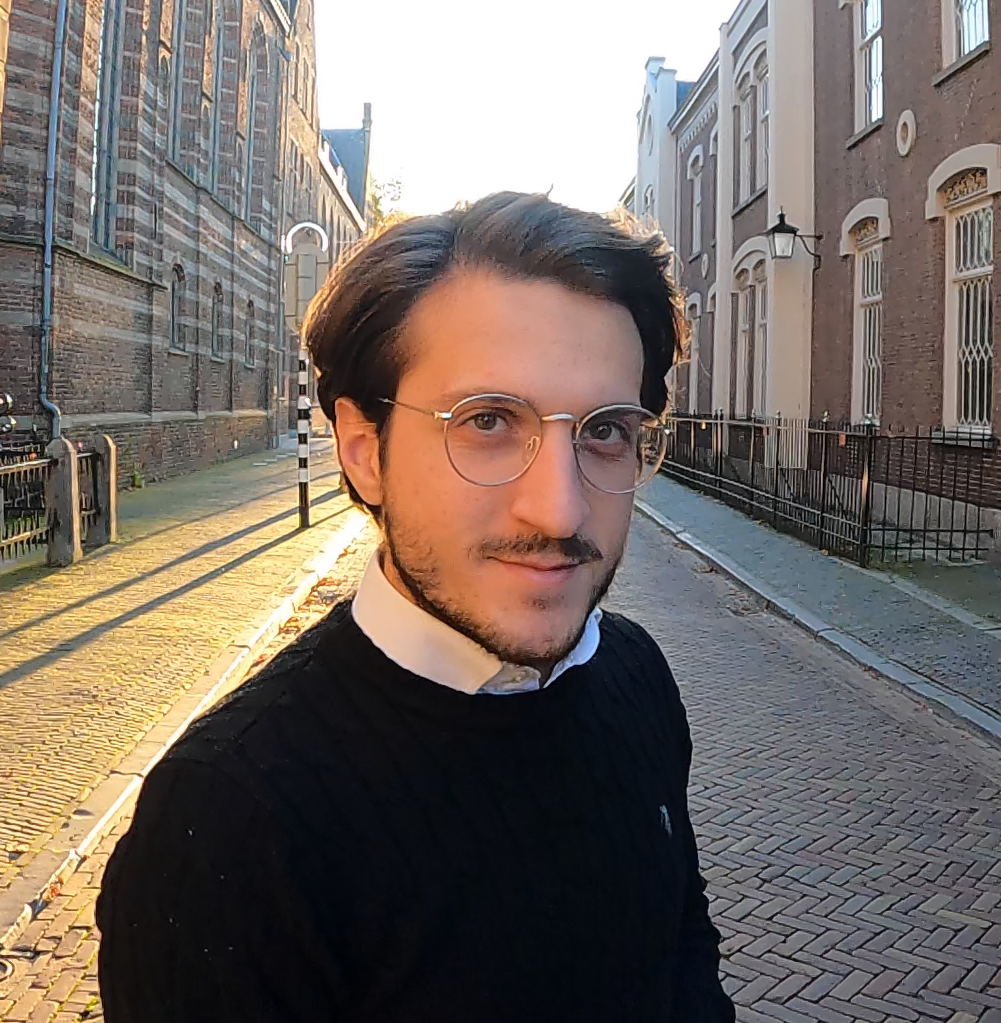}}]{Giovanni Franzese}
	is a PhD Student in the department of Cognitive Robotics at TU Delft, Netherlands since 2019. He received a BSc degree (2016) in Mechanical Engineering and an MSc degree (2018) in Mechatronics and Robotics at Politecnico di Milano, Italy. Since 2022, he is an ELLIS member for artificial intelligence.  
\end{IEEEbiography}
\vspace{-2em} 
\begin{IEEEbiography}
	[{\includegraphics[width=1in,height=1.25in,clip,keepaspectratio]{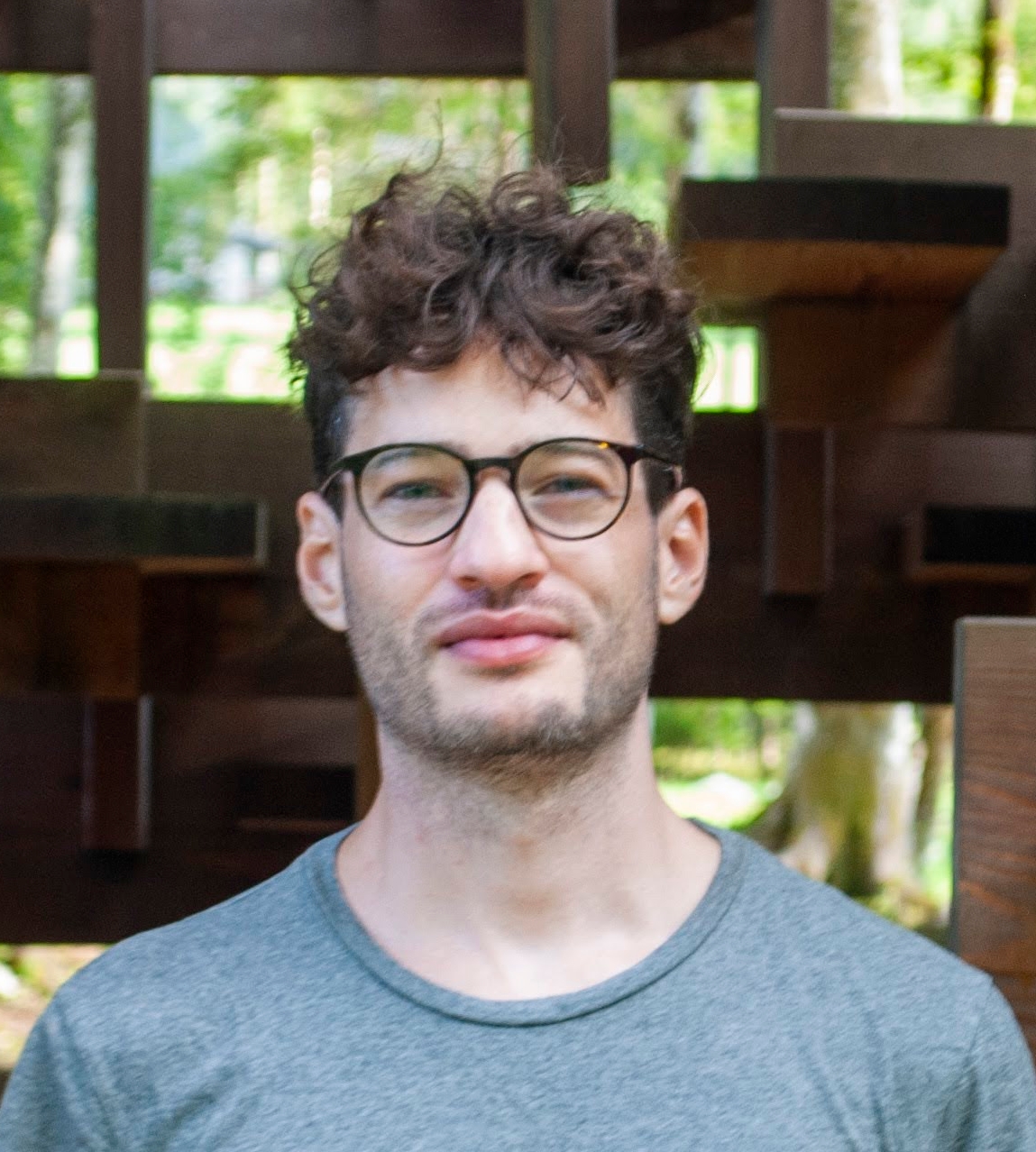}}]{Leandro de Souza Rosa}
	has a BSc (2013) in Computer Engineering and obtained the Ph.D. in 2019 by the Institute of Mathematics and Computer Sciences at The University of São Paulo, Brazil, which was partially developed at the Department of Electrical and Electronic Engineering, Imperial College London, U.K. He worked as a postdoc researcher at the Istituto Italiano di Tecnologia, Genova, Italy, and is currently with the Faculty of Mechanical, Maritime and Materials Engineering, Delft University of Technology, Netherlands.
\end{IEEEbiography}
\vspace{-2em} 
\begin{IEEEbiography}
	[{\includegraphics[width=1in,height=1.25in,clip,keepaspectratio]{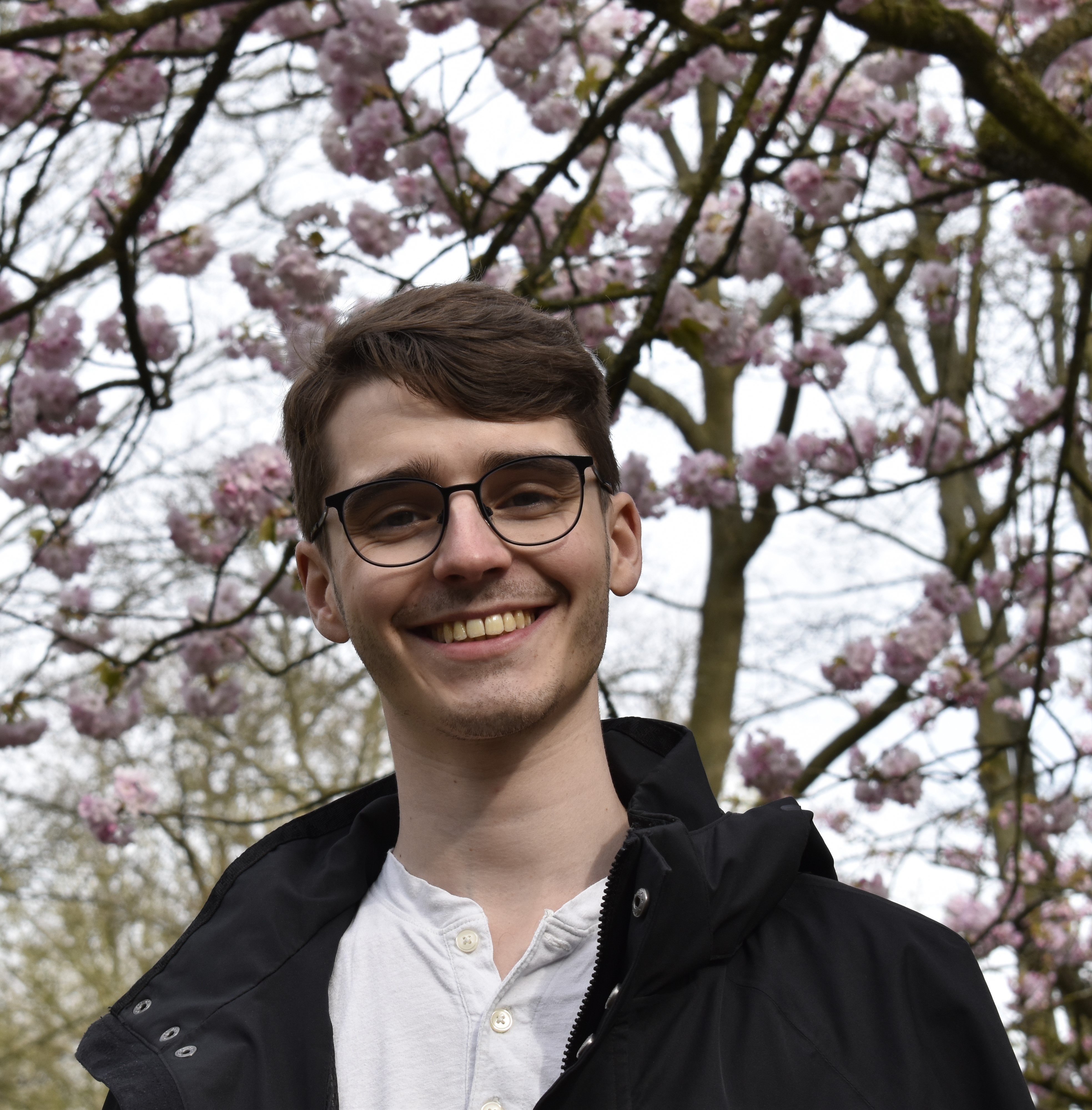}}]{Tim Verburg} is an MSc student in Robotics at the Delft University of Technology. He received his BSc (2021) in Mechanical Engineering at TU Delft, NL with a minor in Robotics. He is interested in mechatronic design and robot learning. 
\end{IEEEbiography}
\vspace{-2em} 
\begin{IEEEbiography}
	[{\includegraphics[width=1in,height=1.25in,clip,keepaspectratio]{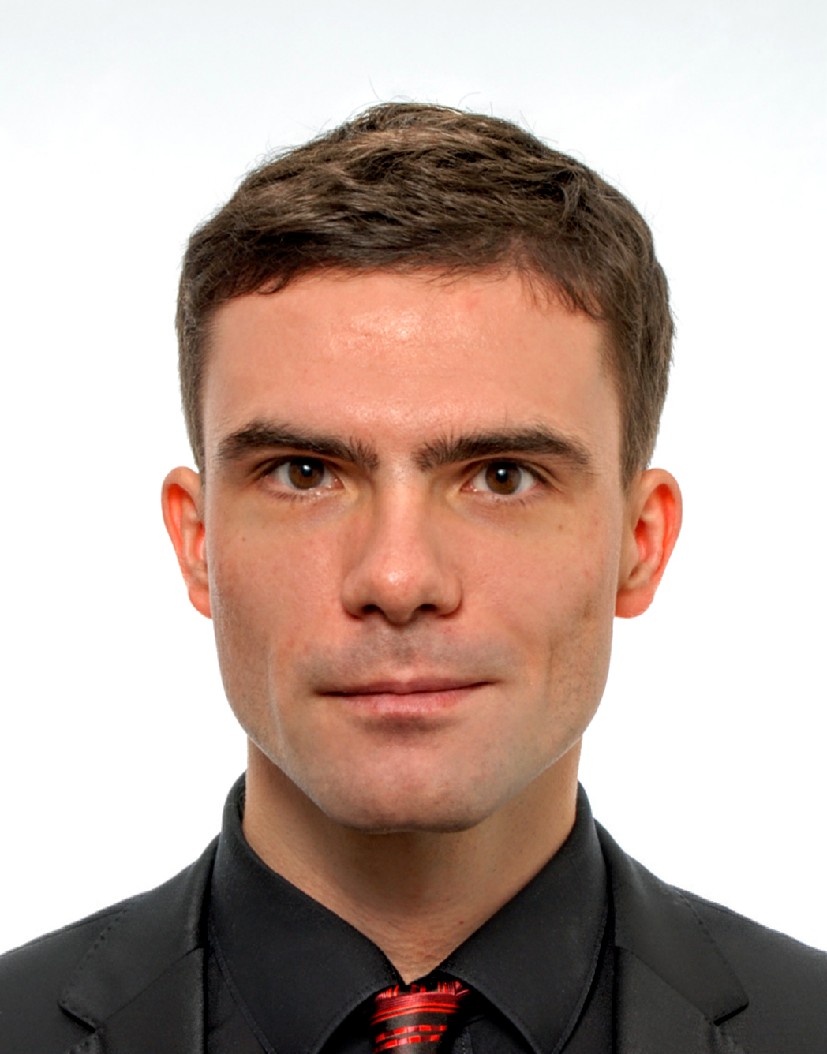}}]{Luka Peternel}
	received a PhD in robotics from University of Ljubljana, Slovenia in 2015. He conducted his PhD studies at the Department for Automation, Biocybernetics and Robotics, Jožef Stefan Institute in Ljubljana from 2011 to 2015, and at the Department of Brain-Robot Interface, ATR Computational Neuroscience Laboratories in Kyoto, Japan in 2013 and 2014. He was with the Human-Robot Interfaces and Physical Interaction Lab, Italian Institute of Technology in Genoa, Italy from 2015 to 2018. Since 2019, he is an Assistant Professor at Cognitive Robotics, TU Delft in the Netherlands.
\end{IEEEbiography} 
\vspace{-2em} 
\begin{IEEEbiography}[{\includegraphics[width=1in,height=1.25in,clip,keepaspectratio]{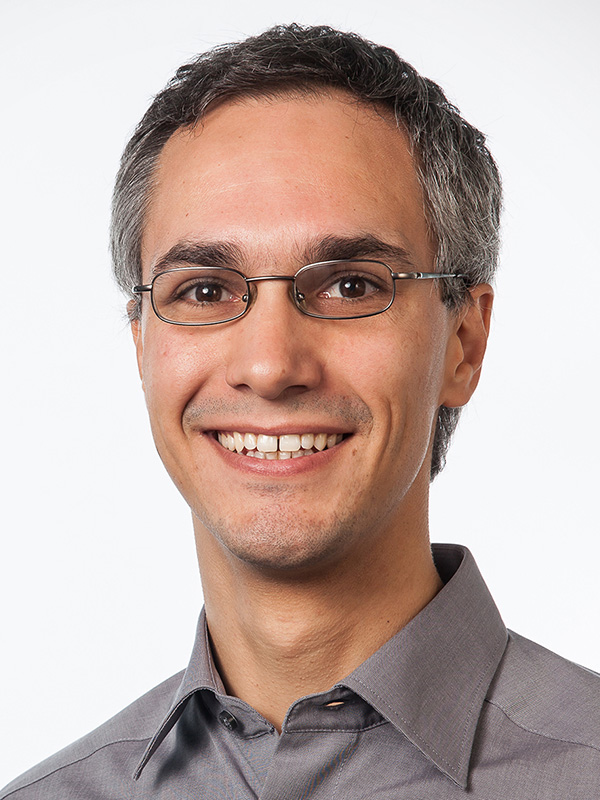}}]{Jens Kober}
	is an associate professor at the TU Delft, Netherlands. He worked as a postdoctoral scholar jointly at the CoR-Lab, Bielefeld University, Germany and at the Honda Research Institute Europe, Germany. He graduated in 2012 with a PhD Degree in Engineering from TU Darmstadt and the MPI for Intelligent Systems. For his research he received the annually awarded Georges Giralt PhD Award for the best PhD thesis in robotics in Europe, the 2018 IEEE RAS Early Academic Career Award, the 2022 RSS Early Career Award, and has received an ERC Starting grant.
\end{IEEEbiography}
\vfill

\end{document}